\documentclass{article}

\usepackage{microtype}
\usepackage{graphicx}
\usepackage{subfigure}
\usepackage{booktabs} 

\usepackage{hyperref}
\usepackage{multirow}

\usepackage[accepted]{icml2024}

\usepackage{amsmath}
\usepackage{amssymb}
\usepackage{mathtools}
\usepackage{amsthm}

\usepackage[capitalize,noabbrev]{cleveref}

\theoremstyle{plain}
\newtheorem{theorem}{Theorem}[section]
\newtheorem{proposition}[theorem]{Proposition}

\theoremstyle{definition}

\theoremstyle{remark}
\newtheorem{remark}[theorem]{Remark}

\usepackage[textsize=tiny]{todonotes}

\icmltitlerunning{Contextual Counting: A Mechanistic Study}

\begin{document}

\twocolumn[
\icmltitle{Contextual Counting: A Mechanistic Study of Transformers \\ on a Quantitative Task}




\begin{icmlauthorlist}
\icmlauthor{Siavash Golkar}{nyu}
\icmlauthor{Alberto Bietti}{fi}
\icmlauthor{Mariel Pettee}{fi,lbnl}
\icmlauthor{Michael Eickenberg}{fi}
\icmlauthor{Miles Cranmer}{uc}
\icmlauthor{Keiya Hirashima}{tok}
\icmlauthor{Geraud Krawezik}{fi}
\icmlauthor{Nicholas Lourie}{nyu}
\icmlauthor{Michael McCabe}{fi,ucb}
\icmlauthor{Rudy Morel}{nyu}
\icmlauthor{Ruben Ohana}{fi}
\icmlauthor{Liam Holden Parker}{fi}
\icmlauthor{Bruno Régaldo-Saint Blancard}{fi}
\icmlauthor{Kyunghyun Cho}{nyu,pd,cifar}
\icmlauthor{Shirley Ho}{fi,nyu,pu}
\end{icmlauthorlist}

\vspace{8pt}
\centerline{\textnormal{The Polymathic AI Collaboration}}

\icmlaffiliation{nyu}{New York University}
\icmlaffiliation{fi}{Flatiron Institute}
\icmlaffiliation{lbnl}{Lawrence Berkeley National Laboratory}
\icmlaffiliation{uc}{University of Cambridge}
\icmlaffiliation{tok}{The University of Tokyo}
\icmlaffiliation{ucb}{University of Colorado Boulder}
\icmlaffiliation{pd}{Prescient Design, Genentech}
\icmlaffiliation{cifar}{CIFAR Fellow}
\icmlaffiliation{pu}{Princeton University}

\icmlcorrespondingauthor{Siavash Golkar}{siavash.golkar@gmail.com}

\icmlkeywords{Machine Learning, ICML}

\vskip 0.3in
]



\printAffiliationsAndNotice{} 

\begin{abstract}
Transformers have revolutionized machine learning across diverse domains, yet understanding their behavior remains crucial, particularly in high-stakes applications. This paper introduces the contextual counting task, a novel toy problem aimed at enhancing our understanding of Transformers in quantitative and scientific contexts. This task requires precise localization and computation within datasets, akin to object detection or region-based scientific analysis. We present theoretical and empirical analysis using both causal and non-causal Transformer architectures, investigating the influence of various positional encodings on performance and interpretability.
In particular, we find that causal attention is much better suited for the task, and that no positional embeddings lead to the best accuracy, though rotary embeddings are competitive and easier to train. We also show that out of distribution performance is tightly linked to which tokens it uses as a bias term.

\end{abstract}

\section{Introduction}
\label{intro}

Transformers have revolutionized the field of machine learning by becoming the backbone of numerous foundation models across various domains, including natural language processing (NLP), computer vision, and scientific research. In NLP, large language models (LLMs)
have set new benchmarks for tasks such as language translation, text generation, and sentiment analysis~\citep{devlin2019bert,brown2020language,touvron2023llama}. In computer vision, Vision Transformers (ViTs) have demonstrated superior performance in image classification and object detection~\cite{dosovitskiy2021image}. Additionally, Transformers are being employed in scientific domains for tasks such as protein folding prediction with models like AlphaFold~\cite{jumper2021highly}, and in chemistry for molecular property prediction~\cite{stokes2020deep}. More recently, Transformer models have been applied to a variety of scientific tasks ranging from foundation models in fluid dynamics to astrophysics and climate science~\cite{Nguyen2023ClimaXAF,McCabe2023MultiplePP,Lanusse2023AstroCLIPCP}.

Understanding how these powerful models make decisions is crucial, as their applications often impact areas requiring high reliability and accountability. Interpretability in machine learning is the field dedicated to making the workings of models more understandable to humans. Broadly, interpretability approaches can be divided into two categories: post-hoc interpretability and mechanistic interpretability. Post-hoc interpretability involves analyzing model outputs after training, using techniques like SHAP values~\cite{lundberg2017unified} and LIME~\cite{ribeiro2016should} to understand feature importance. Mechanistic interpretability, on the other hand, aims to understand the internal workings of the model during its decision-making process. Techniques here include probing model activations, studying attention patterns, and analyzing the learned representations within the model~\cite{clark2019bert,vig2019multiscale}.

In the context of NLP, various toy problems such as the induction head and recall tasks have been extensively studied to probe specific capabilities of Transformers. For example, induction heads are known to play a key role in pattern recognition tasks, where the model learns to identify and generalize sequences of tokens~\cite{olah2020zoom}. However, these tasks are predominantly tailored to NLP applications. There is a significant need for similar interpretability challenges that cater to quantitative models used in scientific computations and data analysis.

In this paper, we introduce a novel toy problem designed to advance the interpretability of Transformer models in quantitative and scientific contexts. This task, which we call \emph{contextual counting}, requires the model to identify a specific region of interest within a dataset and perform accurate counting. It simulates scenarios where precise localization and subsequent computation are critical, such as in object detection or region-based analysis in scientific data. We provide comprehensive theoretical and empirical studies utilizing both causal and non-causal Transformer architectures. We explore the impact of positional encodings on the performance of these tasks, offering insights into how different positional information influences model interpretability and performance in quantitative settings. Training a variety of small encoder-decoder models, we find that despite the absence of a specific causal structure in the problem, causal Transformers perform far better than non-causal ones. We identify the circuitry used in the best performing models and find that RoPE is much more likely to find good solutions, NoPE is second best in this respect and Alibi and Absolute position provide poor performance. We also show explicitly how generalizability to out of distribution domains can be traced to the use of different tokens as bias terms.

For an overview of the different types of Transformer configurations explored in this work see Appendix~\ref{app:configs}.

\subsection*{Related Work}
\label{sec:related_work}
Understanding the internal mechanisms of Transformers, particularly in language modeling, has recently been an active area of research, from dissecting pre-trained large language models~\citep{elhage2021mathematical,meng2022locating,wang2022interpretability,geva2023dissecting} to studying how simple mechanisms arise in controlled settings~\cite{zhang2022unveiling,nanda2023progress,liu2023Transformers,li2023Transformers,bietti2023birth,allen2023physics}.
The empirical limitations of Transformers on certain quantitative tasks has been observed in several works~\citep{dziri2024faith,jelassi2023length,Golkar2023xValAC}. While some ad-hoc changes may improve performance in specific settings, such as special prompt formatting, or chain-of-thought reasoning~\citep{jelassi2023length,lee2023teaching,dziri2024faith}, trying to achieve perfect accuracy on numerical tasks may require learning to simulate intricate algorithms that are beyond the expressive capabilities of basic Transformers~\citep{hahn2020theoretical,merrill2022saturated,sanford2024Transformers}.
Our work instead highlights how Transformers may approximate numerical solutions by learning mechanisms that approximate continuous computations, while also leveraging discrete operations such as selective attention to specific regions of the input.



\section{Task Description}
\label{sec:counting_task}
In this section, we introduce the Contextual Counting task and present some theoretical analyses regarding what kind of models can/cannot tackle this task efficiently.

\subsection{Problem statement}
In this task, the input is a sequence comprised of zeros, ones, and the square bracket delimiters: \{0, 1, [, ]\}. Each sample is a sequences of ones and zeros, with a number of regions marked with the delimiter tokens. The task is then to count the number of ones that appear within the delimited region. Equation~\eqref{eq:counting_example} gives an example with sequence length 16 and three regions. 
\begin{align}
    \label{eq:counting_example}
    &\text{input} = \texttt{[ 0 ] [ 1 0 1 ] 0 [ 1 ] 1 [ ] 0}\notag\\
    &\text{target} = [0, 2, 1, 0]
\end{align}
For simplicity we take the regions to be non-overlapping.

In order to extract the target values using a Transformer architecture, we use an encoder-decoder setup and provide the decoder with a fixed prompt comprised of the label of the regions. In the example~\eqref{eq:counting_example} above, the prompt would be:
\begin{equation}
    \label{eq:counting_prompt}
    \text{prompt} = [0, 1, 2, 3]
\end{equation}

In this work, for our empirical examples we fix the number of regions to 4 and the sequence length to 512. By fixing the number of regions and sequence length, we can explore how the solutions found in various settings generalize to unseen number of regions and different sequence lengths. 

\paragraph{Relevance:} More than an instructive example in mechanistic understanding of Transformers, this task emulates quantitative problems where precise sensitivity to regional boundaries is required. An example of this is calculations of properties inside detected objects (e.g. counting the number of specific neuro-receptors inside a specific neuron).

Furthermore, this task cannot be solved with current state of the art large language models. For examples of how some of the current SOTA LLMs respond to this task, see Section~\ref{sub:llm} and Appendix~\ref{app:llms_fail}.

\subsection{Theoretical insights}

In this section, we provide some theoretical insights on the problem. Specifically, we show that a Transformer with one causal encoding layer and one decoding layer - one attention head each - and NoPE can solve the contextual counting task for arbitrary sequence length and number of regions. We also provide some results regarding the challenge of solving this task with non-causal Transformers. The proof of the statements in this section can be found in Appendix section~\ref{app:proofs}.

In order to construct this solution, we first introduce the concept of \emph{contextual position} (CP). In short, contextual position refers to positional information in a sequence which - as opposed to global or relative position code - is only meaningful in the context of the problem. For example, in the current problem, a contextual position information is the region number. In an example with three regions we would have:
\begin{align*}
    \label{eq:contextual_position}
    \text{input} =\:& \texttt{0 1 1 [ 1 0 1 ] 0 [ 1 ] 1 [ ] 0}\notag\\
    \text{CP} =\:& \texttt{- - - 1 1 1 1 1 - 2 2 2 - 3 3 -}
\end{align*}
This extra piece of information - which for the contextual counting task we call the regional contextual position - helps to disambiguate the different regions of the sequence based on the context.

\begin{proposition}
\label{prop:contextual-position}
(informal) If the regional contextual position information is linearly decodable from the latent representation of the tokens at some layer of a Transformer, the Contextual Counting task can be solved with a single additional layer. 
\end{proposition}
Specifically, an attention layer can use this information to attend to a unique region and infer the number of ones by summing value vectors coming from the 1-tokens in this region, similar to the construction used in~\citet{Kazemnejad2023TheIO}. 

\begin{proposition}
\label{prop:contextual-position}
(informal) A causal Transformer with a single layer and no position encoding (NoPE) can infer the regional contextual position for arbitrary number of regions and arbitrary sequence lengths.
\end{proposition}

The combination of the previous two propositions imply that a two layer causal Transformer with NoPE can solve the Contextual Counting task. We will see explicitly that in the best performing models, the first layer extracts the  regional contextual position and the second layer attends uniquely to the appropriate region as in the above two propositions.

For Transformers that are non-causal (a.k.a.~bidirectional), the task is more complicated. The following two propositions outline some of the difficulty in this case.

\begin{proposition}
    \label{prop:nope-noncausal}
    (informal) A non-causal Transformer with no position code and a permutation invariant output head cannot solve the Contextual Counting task.
\end{proposition}
This is because such a Transformer is fully permutation invariant and the Contextual Counting task is sensitive to the ordering of the sequence elements. Examples of permutation invariant output heads are averaging over the tokens or  using a cross-attention head as in a decoder.

\begin{proposition}
    \label{noncausal-abs}
    (informal) To exactly emulate a causal attention profile, a non-causal attention layer with Absolute Position code would need an embedding space that has dimension at least as large as the length of the sequence.  
\end{proposition}
In other words, using AbsPE to produce a causal attention map on a sequence of $T$ tokens, the embedding space needs to be at least $T$ dimensional.

\begin{remark}
    \label{noncausal-rel}
    As RoPE and Alibi are constructed to have attention profiles that are only sensitive to distance between the tokens, their extensions to non-causal Transformers leads to attention profiles that cannot distinguish between the case where a key token comes before the query token and the case where the same token comes after query token. Therefore, a non-causal Transformer with RoPE or Alibi cannot emulate a causal Transformer.  
\end{remark}

The above propositions and remark imply that the results pertaining to causal Transformers, cannot be readily applied to non-causal Transformers. And as we will see, causal Transformers outperform non-causal ones significantly. 





\section{Experimental Results}
\label{sec:experiments}
\begin{figure}
    \centering
    \includegraphics[width=\columnwidth]{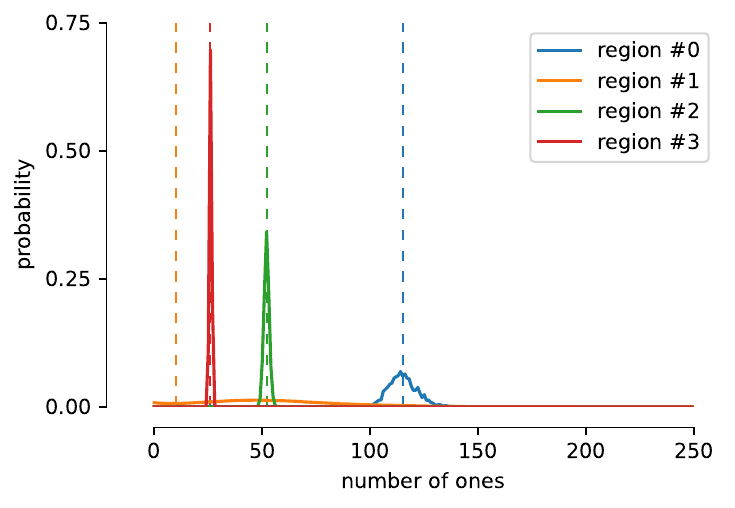}
    \vspace{-18pt}
    \caption{Typical output of a trained model on the Contextual Counting task. The model outputs probability distribution over the number of ones in the relevant region. In this case, regions 0, 2, and 3 predict the correct numbers (given by the dashed line) but region 1 has failed to learn.}
    \label{fig:counting_output}
\end{figure}

In this section we describe the results of training various Transformer architectures on the Contextual Counting task. We trained encoder-decoder models where the encoder and the decoder are each comprised of a single Transformer block each with one attention head. The output of the model is 4 vectors coming from the 4 prompt tokens fed into the decoder, one for each region (Eq.~\ref{eq:counting_prompt}). Each vector encodes the probability distribution over the number of ones that the model believes exist in the relevant region. An example of the output of a trained model is given in Fig.~\ref{fig:counting_output}. Note that the model can implement different algorithms for the different region, for example in Fig.~\ref{fig:counting_output}, the model has failed to learn the correct dependence for region~1 but is correctly predicting this number for other regions. Because of this, in what follows we report the performance of the different regions separately.

To get an intuition for the circuitry in the models, we trained Transformers with the decoder comprised of only a single cross-attention layer, i.e. no MLP layer and no self-attention in their decoder block. In this case, the output of the model becomes bilinear as a function of the output of the decoder layer, making interpretation simpler. We look at the effect of the MLP in Appendix~\ref{app:MLP}.

For each configuration we train three models with different random seeds. As we report the different regions separately, this leads to 12 individual points for each configuration.

\subsection{Training results}

\begin{figure}
    \centering
    \includegraphics[width=\columnwidth]{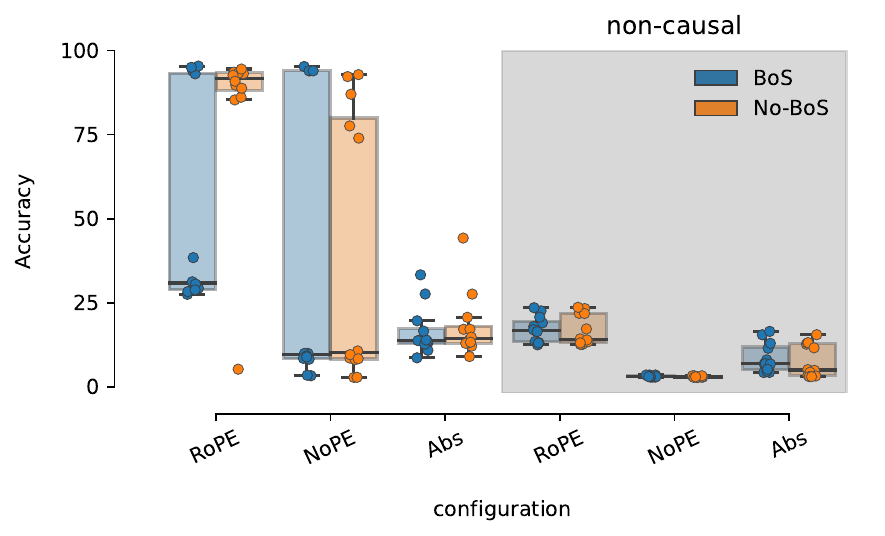}
    \vspace{-18pt}
    \caption{Model Accuracy on the Contextual Counting task.  Here the input sequence is length 512 with 4 regions. The results in the shaded region denote models trained with non-causal attention. RoPE and NoPE outperfrom absolute position code and non-causal models fail to learn. For Alibi see Sec.~\ref{app:alibi}.}
    \label{fig:counting_acc_orig}
\end{figure}

The results of the training with various configurations on an independent test set is given in Fig.~\ref{fig:counting_acc_orig}. We highlight a number of features in these results. First, there are models that find solutions with close to 100\% top-1-accuracy and exactly 100\% top-5 accuracy. Second, non-causal models achieve very low performance on this task, reflecting the difficulty of the task for non-causal models as mentioned in the previous section. Third, NoPE performs much better than AbsPE.  We will  come back to these points as we discuss the model circuitry below. Alibi results are discussed below and in Sec.~\ref{app:alibi}.

\label{sec:counting_resutls}

\begin{figure}
    \centering
    \includegraphics[width=\columnwidth]{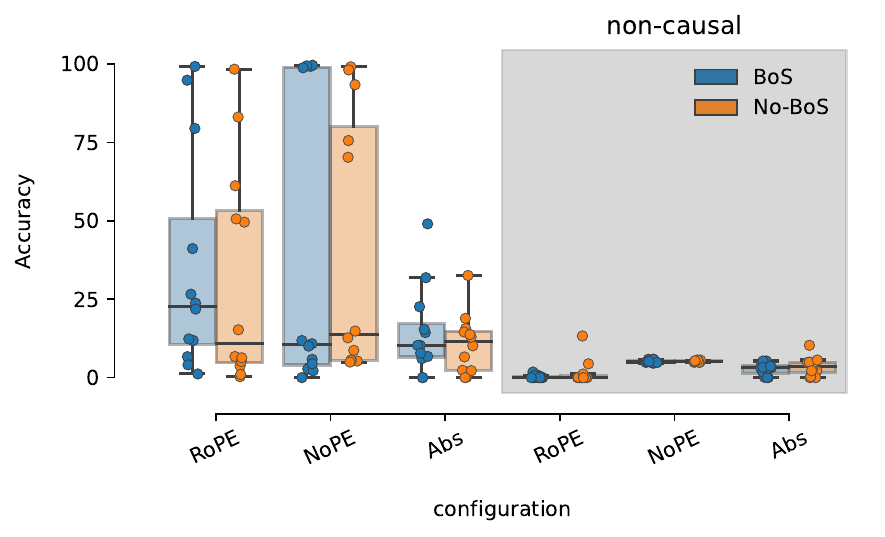}
    \vspace{-18pt}
    \caption{Generalization performance on test samples with shorter sequences (T=300). Of the models that perform well in-distribution, only a few generalize to shorter sequence lengths.}
    \label{fig:counting_acc_shorter}
\end{figure}

\subsection{Generalization}

To understand the nature of the solutions found, it is informative look at whether or not each solution generalizes to changes in the statistics of the input. 

\paragraph{Different Sequence Length.} In the training set, the length of the input sequence is always fixed at 512. Here we look at the performance of the trained models when the sequence length is shortened.\footnote{We look at generalization to shorter sequences instead of longer sequences for two reasons. First, as we will see, this is not trivial in this task. Second, generalization to longer sequences is simply not possible for AbsPE.} This can be seen in Fig.~\ref{fig:counting_acc_shorter}. Comparing this figure to Fig.~\ref{fig:counting_acc_orig}, we see two classes of solutions. One where the performance stays high and another where the performance drops to chance level. 

\begin{figure}
    \centering
    \includegraphics[width=\columnwidth]{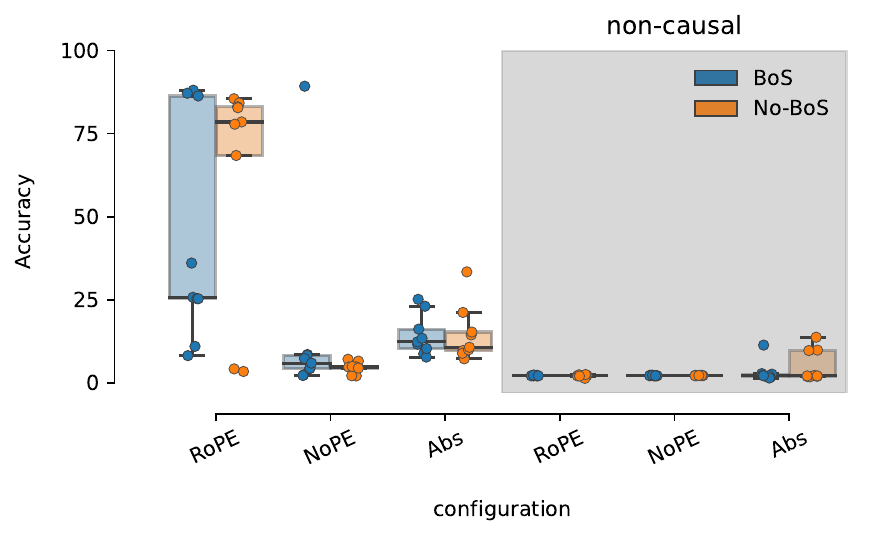}
    \vspace{-18pt}
    \caption{Generalization performance on test samples with three regions. In this case RoPE generalizes much better than NoPE. }
    \label{fig:counting_acc_3bounds}
\end{figure}

\paragraph{Different Number of Regions.} We also look at changing the input distribution by changing the number of regions in the input. Specifically, instead of 4 regions we have 3 indicated by 3 sets of square brackets. The performance in this setting can be seen in Fig.~\ref{fig:counting_acc_3bounds}. Again we see that some models retain their performance while others drop to chance level.

\begin{figure*}[ht]
    \centering
    \includegraphics[width=\textwidth]{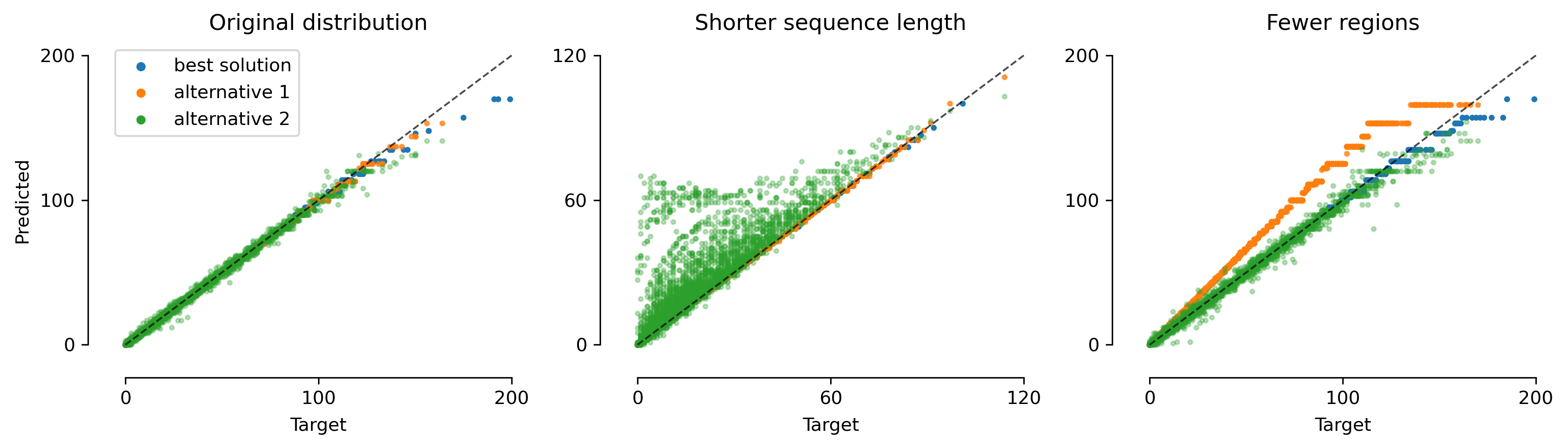}
    \vspace{-18pt}
    \caption{Prediction of three different solutions in the original distribution as well as shorter sequences and fewer number of regions. The various solutions types suffer from different failure modes when evaluated on out-of-distribution samples. The difference in behavior can be traced to the use of different tokens as biasing terms.}
    \label{fig:counting_loss_nomlp_varsols}
\end{figure*}

Figure~\ref{fig:counting_loss_nomlp_varsols} shows the behavior of 3 different types of solutions when evaluated on the out-of-distribution test samples mentioned above. All three solutions achieve good performance in-distribution (Fig.~\ref{fig:counting_loss_nomlp_varsols}-left). However, while one solution maintains its performance on the generalization task (blue dots), one solution generalizes to shorter sequences but fails on fewer regions (orange dots), and another generalizes to fewer regions but fails on shorter sequences (green dots). 
We see that even though the three solutions perform similarly on the original distribution, their generalization performance is different.

\subsection{Learned circuits}

In this section, we look at the trained Transformers and describe in detail how they solve the Contextual Counting task. We start by looking at the model that attains the best performance on both the original distribution as well as in the generalization experiments (top-most point in Figs~\ref{fig:counting_acc_orig}, \ref{fig:counting_acc_shorter}, \ref{fig:counting_acc_3bounds} and the solution indicated as best in Fig~\ref{fig:counting_loss_nomlp_varsols}). This is a causal model with NoPE and a BoS token. 

\begin{figure}
    \centering
    \includegraphics[width=0.9\columnwidth]{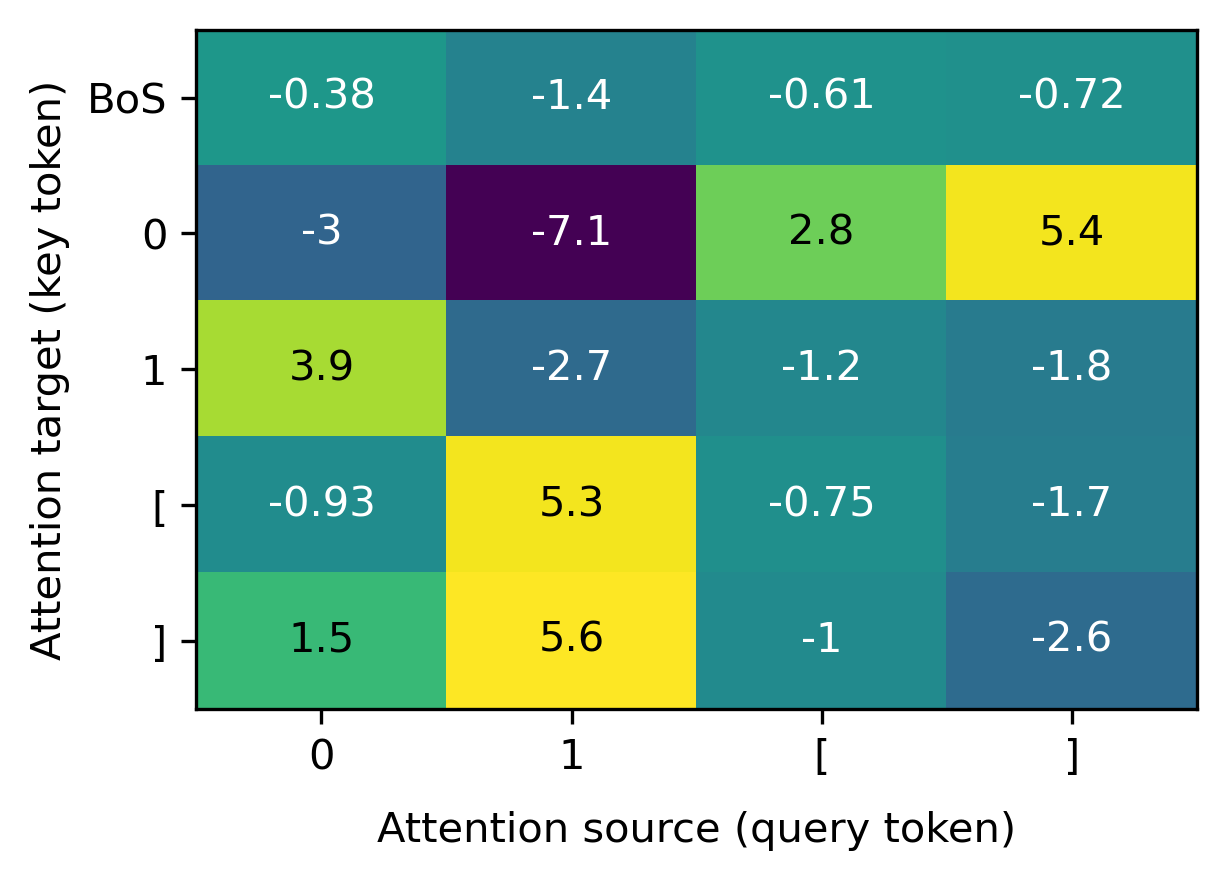}
    \vspace{-4pt}
    \caption{The encoder attention pattern of the best model. The numbers show the pre-softmax attention weights. We see that the 1-tokens only attend to the delimiters. This allows the 1 tokens to calculate the contextual position.}
    \label{fig:counting_loss_nomlp_bestsol_enc_att}
\end{figure}

\paragraph{Encoder.} As the model has no positional encoding, the self-attention pattern of the encoder is only determined based on context. Furthermore, since this is the first layer, the context is only dependent on the identity of the token. We can therefore get a complete picture of the attention pattern by looking at the token by token attention matrix given in Fig.~\ref{fig:counting_loss_nomlp_bestsol_enc_att}.

The most important feature of this attention pattern is that the 1-tokens attend only to the preceding delimiters. In this way, they can tag themselves with their  contextual position.  We can verify this explicitly by looking at the PCA of the latent representation of the 1-tokens after the self-attention layer. Since the 1-tokens only attend to the delimiters, their latent representation is rank two. We can therefore get an accurate representation of their information content by looking at their 2D PCA projection. Following the formulae for self-attention, we can simply verify that this construction leads to a latent state that can differentiate the different regions of the 1-tokens, as in Prop.~\ref{prop:contextual-position}.

\begin{figure}
    \centering
    \includegraphics[width=0.95\columnwidth]{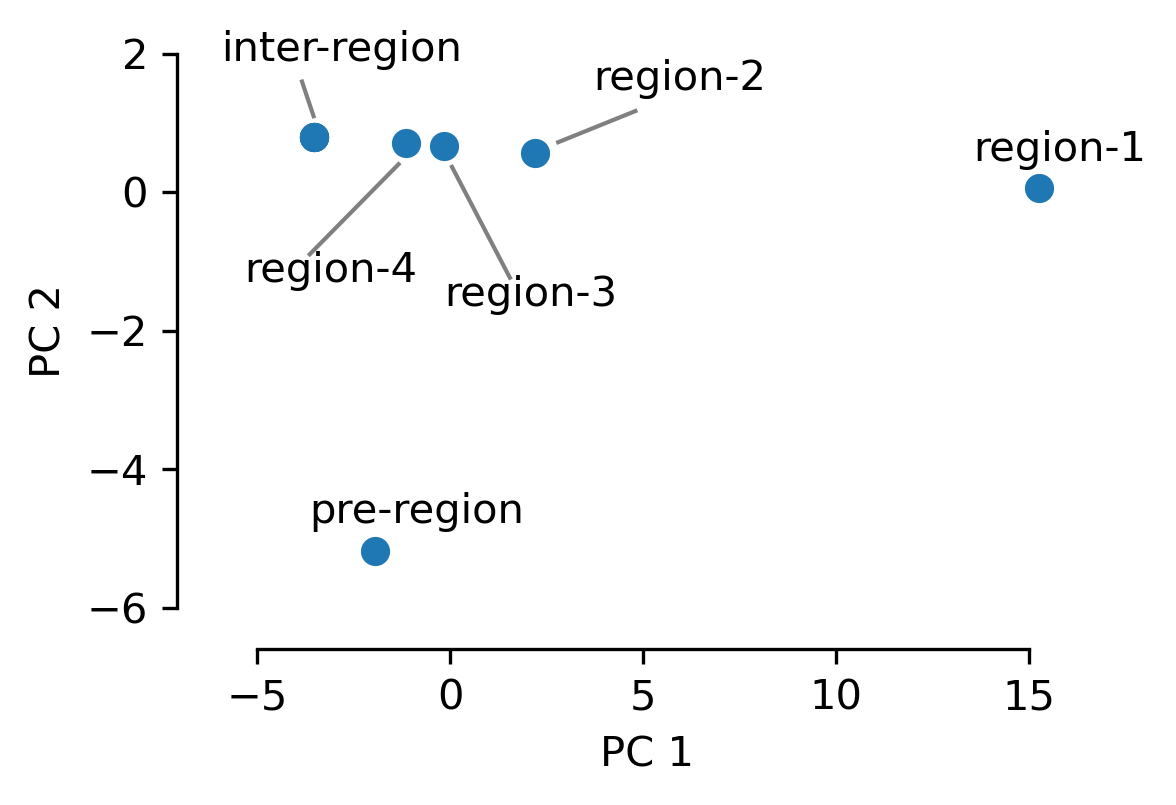}
    \vspace{-4pt}
    \caption{Projection of the output of the self-attention layer for the 1-tokens in the different regions of the input. We see that the regional contextual position is encoded in the top principle component of the 1-token value vectors. Here pre-region and inter-region refer to parts of the sequence that are respectively befoer the first delimiter and in between the regions of interest.}
    \label{fig:counting_loss_nomlp_bestsol_ones_postatt}
\end{figure}

Figure~\ref{fig:counting_loss_nomlp_bestsol_ones_postatt} shows the projection of the output of the self-attention layer on the 1-tokens in the different regions. We see that the representation seperates out the different parts of the sequence accurately. This mechanism of tagging the contextual location of the 1-tokens is sufficient for explaining the behavior of the encoder.

\begin{figure}
    \centering
    \includegraphics[width=0.95\columnwidth]{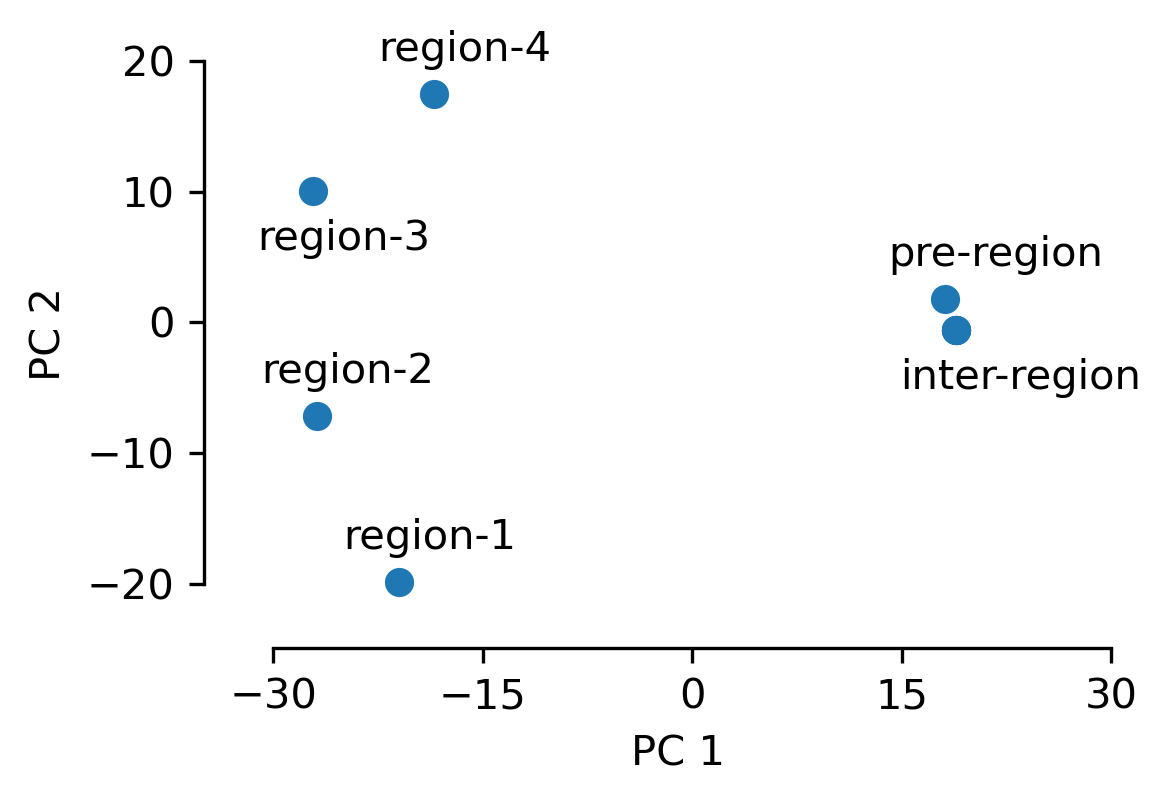}
    \vspace{-4pt}
    \caption{Projection of the keys of the cross-attention layer for the 1-tokens in the different regions of the input. The regional contextual position is now spread out and separated from the inter-region.}
    \label{fig:counting_loss_nomlp_bestsol_keys}
\end{figure}

\paragraph{Decoder.} To understand the mechanisms at play in the decoder, we need to look at the decoder cross-attention attention map as well as the values. Figure~\ref{fig:counting_loss_nomlp_bestsol_crossatt} shows the attention pattern of the cross-attention layer of the decoder for a generic example. We see that the cross-attention head only attends to the 1-tokens in the relevant region as well as to the BoS-token. As we saw in the encoder, the 1-tokens tag themselves with the relevant contextual position information such that the decoder can attend to them and not to anything else. This can also be seen in the keys of the cross-attention module in the decoder (Fig.~\ref{fig:counting_loss_nomlp_bestsol_keys}).

\begin{figure}
    \centering
    \includegraphics[width=0.95\columnwidth]{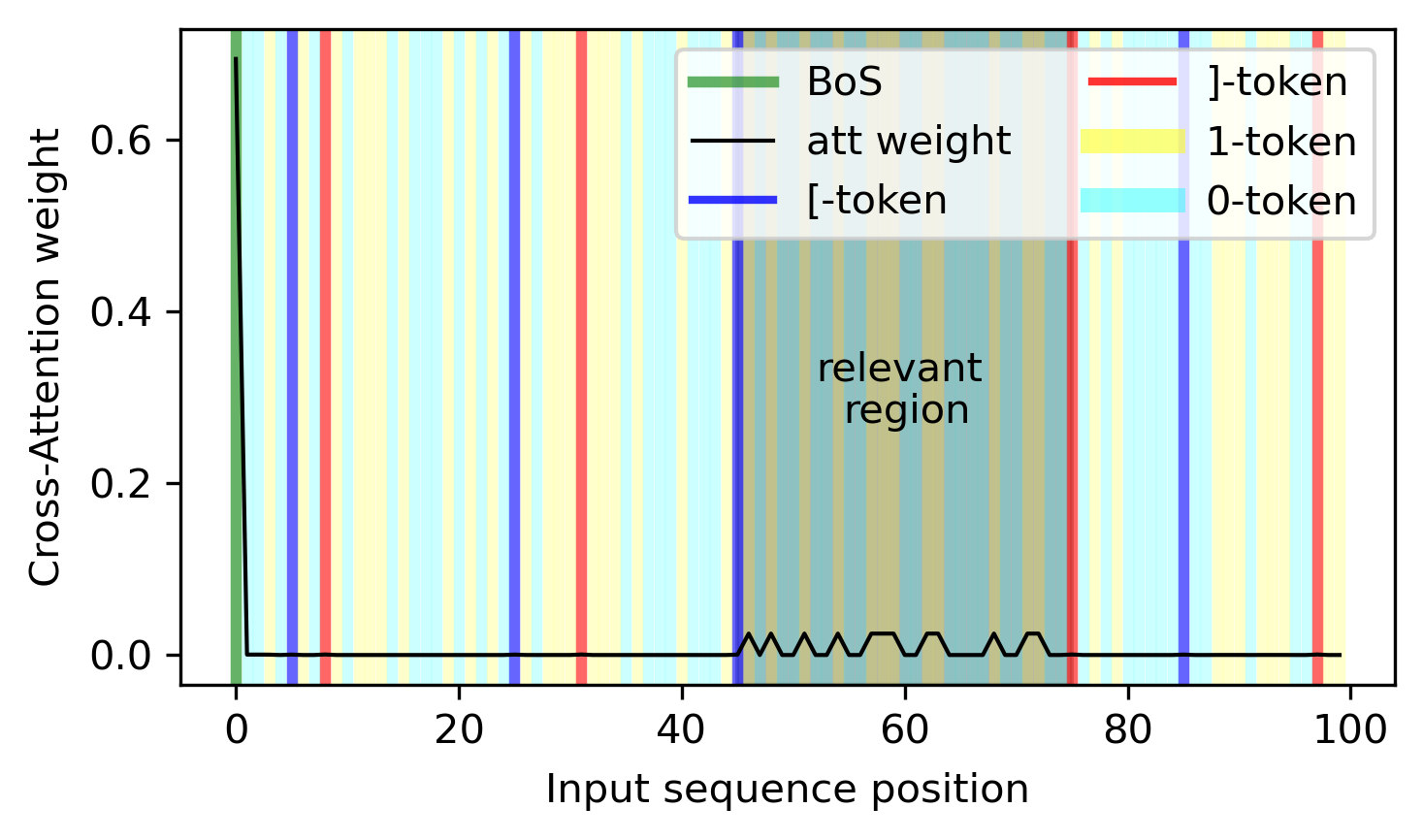}
    \vspace{-4pt}
    \caption{The attention weights of the cross-attention module of the decoder with the query token representing the third region (shaded). Different colors denote the identity of the token at the specific position. As with other cases, the model was trained on length 512 input sequences but for demonstration, we show a sample of length 100.}
    \label{fig:counting_loss_nomlp_bestsol_crossatt}
\end{figure}

\begin{figure}
    \centering
    \includegraphics[width=0.95\columnwidth]{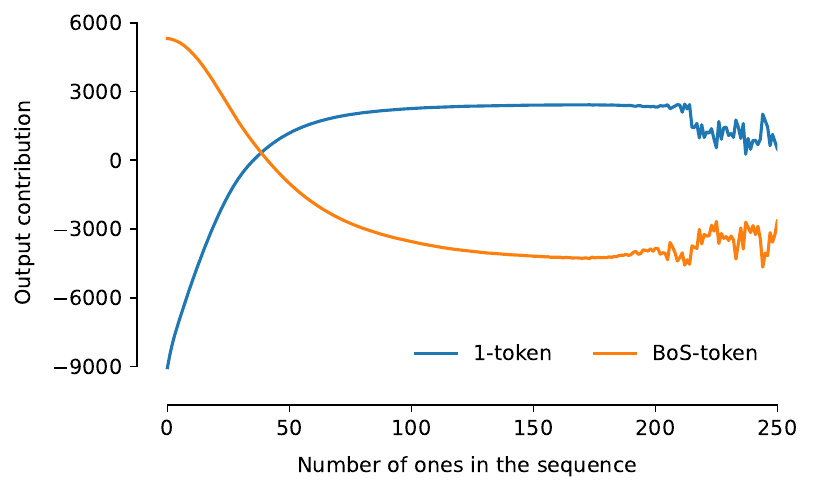}
    \vspace{-4pt}
    \caption{Contribution of the different tokens to the output of the model. The model uses a linear sum of these two lines to construct the final output of the network (Eq.~\ref{eq:bestsol_formula}.)}
    \label{fig:counting_loss_nomlp_bestsol_val_contribs}
\end{figure}

Finally, we look at the values of the cross-attention module. As we do not have an MLP layer in these models, the output of the model (which after a softmax is shown in Fig.~\ref{fig:counting_output}) is a linear sum of the values in this layer. Since the cross-attention module only attends to the BoS and 1-tokens, we need only look at the value vectors associated to 1-tokens and BoS. Figure~\ref{fig:counting_loss_nomlp_bestsol_val_contribs} shows these effective values (i.e. the values after being multiplied them with the linear layers that follow the cross-attention module).

\begin{figure}
    \centering
    \includegraphics[width=0.95\columnwidth]{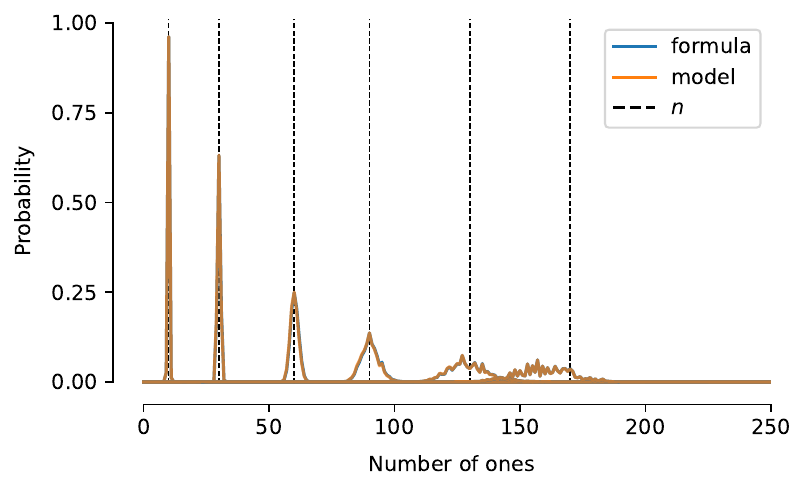}
    \vspace{-4pt}
    \caption{Comparison of the function in Eq.~\eqref{eq:bestsol_formula} vs. the output of the Transformer for different number of 1-tokens in the relevant region. The formula which analytically captures the dependence on the number of 1-tokens $n$ in the relevant region, accurately matches the output of the model.}
    \label{fig:counting_loss_nomlp_bestsol_probs}
\end{figure}

With the above in place, we see that the model is implementing the following rather simple formula for the output of the model as a function of $n$ number of ones in the relevant region:
\begin{equation}
    \label{eq:bestsol_formula}
    f(n) = \text{softmax} \big( n a(\texttt{1}) v_{e\!f\!f}(\texttt{1}) + a(\texttt{BoS}) v_{e\!f\!f}(\texttt{BoS})\big), 
\end{equation}
where $a(\texttt{1})$ and $a(\texttt{BoS})$ are the normalized attention weights and $ v_{e\!f\!f}$ are the value vectors multiplied by the linear layers that follow the cross-attention layer (see Fig.~\ref{fig:counting_loss_nomlp_bestsol_val_contribs}). 
The $n$ dependence of this formula\footnote{The attention weights themselves have an implicit dependence on $n$ because of the softmax normalization.} is coming from the fact that the cross-attention module is only attending the the 1-tokens in the appropriate  region. The model is then balancing the increasing and decreasing curves given in Fig.~\ref{fig:counting_loss_nomlp_bestsol_val_contribs} to arrive at a profile that peaks at $n$.  Figure~\ref{fig:counting_loss_nomlp_bestsol_probs} show the probability associated with this function for various values of $n$. For comparison, in Fig.~\ref{fig:counting_loss_nomlp_bestsol_probs} we have also included the output of the model which shows that the formula is a good approximation of the algorithm implemented in the model.


Note that in this solution, even though the number of ones $n$ is known, the output of the model is not sharply centered at $n$. Indeed as $n$ grows, the width of the envelope predicted by the model grows as well. This is a side-effect of this specific solution. 

Finally, we point out that in this circuit, it is clear that the role of the BoS-token is that of a bias term. That is, it provides a constant addition to the output of the model. Without this term, the model would only attend to the 1-tokens, and the $n$ dependence of the solution which is crucial for our task, would go away.  The use of a bias token as a necessary component of a Transformer circuit implementing counting was previously seen in~\cite{Kazemnejad2023TheIO}.

\subsection{Why do some solutions not generalize?}

\begin{figure}
    \centering
    \includegraphics[width=0.95\columnwidth]{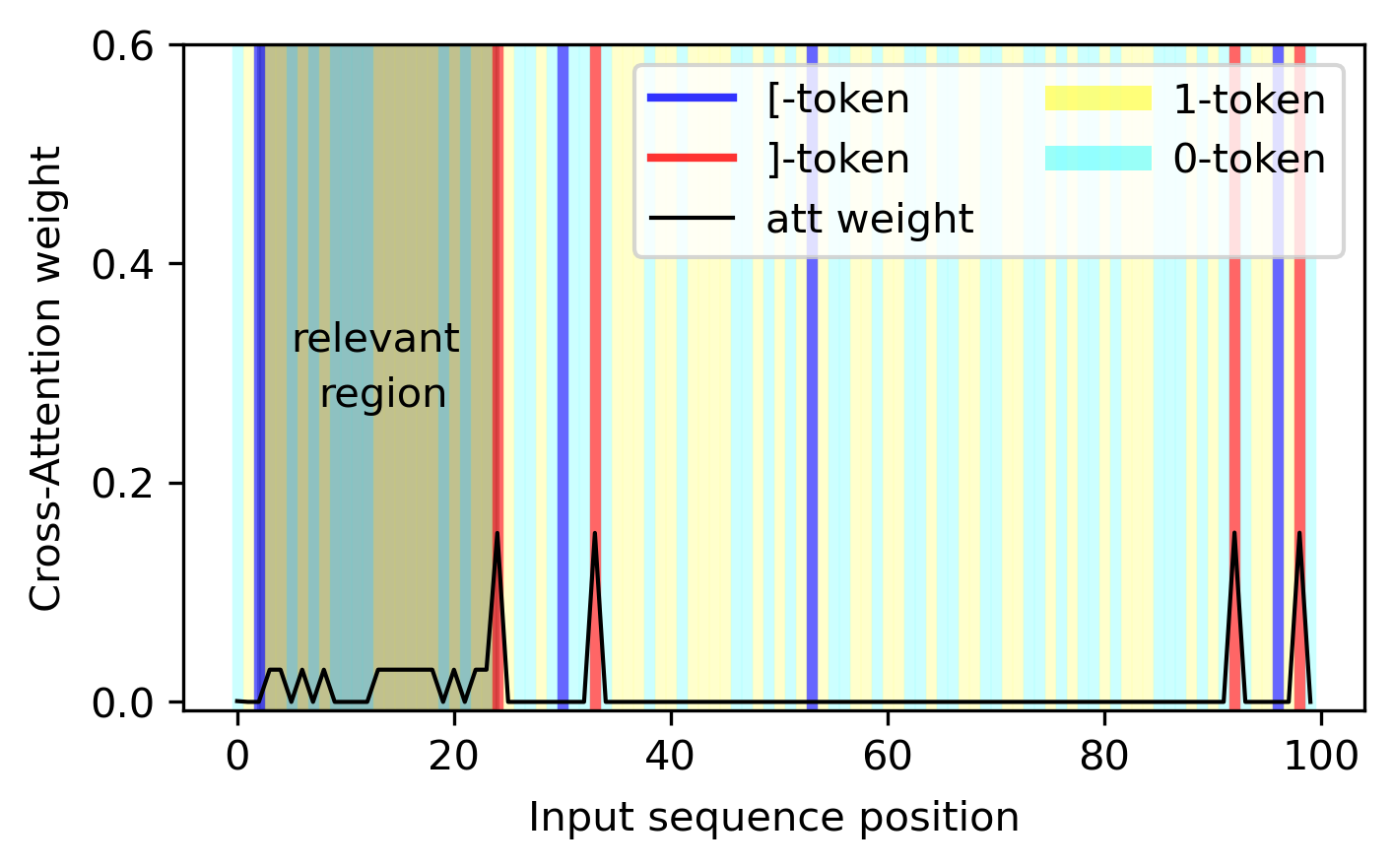}
    \vspace{-4pt}
    \caption{The attention weights of the cross-attention module of the decoder in the model without BoS-token. The shaded regions denote the identity of the tokens at the indicated position. As with other cases, the model was trained on length 512 input sequences but for demonstration, we show a sample of length 100.}
    \label{fig:counting_loss_nomlp_bestsol_crossatt_nobos}
\end{figure}

In this section we describe why some models with good in-distribution performance fail to generalize in the out-of-distribution cases.

An instructive example to consider is the highest accuracy model with NoPE and no BoS-token. The behavior of the encoder in this model is very similar to that of the model with BoS with the 1-tokens tagging their own position by attending to the delimiter tokens. The main difference is that in the decoder, the attention weights are now focused on 1-tokens and the ]-tokens (Fig.~\ref{fig:counting_loss_nomlp_bestsol_crossatt_nobos}).

\begin{figure}
    \centering
    \includegraphics[width=0.95\columnwidth]{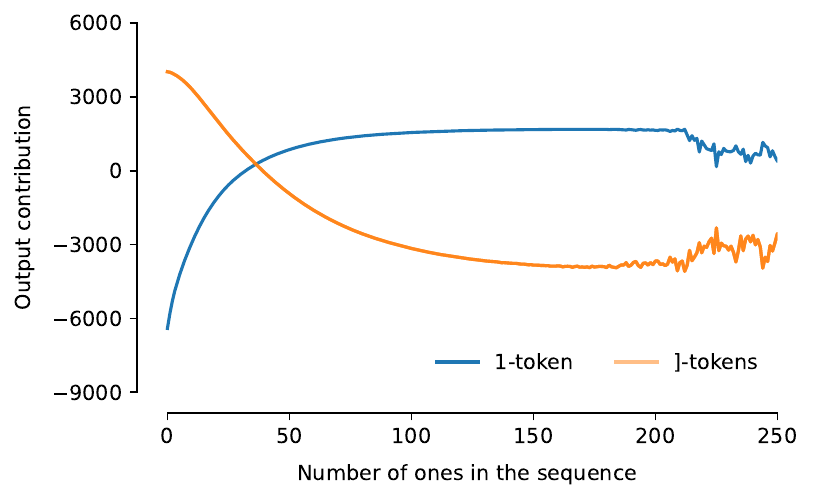}
    \vspace{-4pt}
    \caption{Contribution of the different tokens to the output of the model with no BoS token. Here the role of the BoS token is taken up by the ]-tokens leading to failure of generalization when the number of regions changes.}
    \label{fig:counting_loss_nomlp_bestsol_val_contribs_nobos}
\end{figure}

Looking at the contribution of the ]-tokens to the output, we see that the values of the ]-tokens are almost identical to that of the BoS-token in the previous case (compare Fig.~\ref{fig:counting_loss_nomlp_bestsol_val_contribs_nobos} to~\ref{fig:counting_loss_nomlp_bestsol_val_contribs}). That is, the role of the bias term, previously played by the BoS-token is now taken up by this delimiter token. This is because the number of delimiters in the training set was always set to 4 and it explains why the model does not generalize when the number of regions is changed. However, this solution \emph{does} generalize to other input sequence lengths.

 Yet another example of the model using spurious constants as a biasing term is a solution where the model uses the 0-token in the place of the BoS-token. Because the input sequence has 0s and 1s drawn from a Bernoulli distribution, the expected value of the total number of 0s in the sequence is equal to half of the sequence length. In some cases, the model uses the number of 0-tokens as a bias and therefore does not generalize to other sequence lengths while generalizing to other number of regions. 
 
 The fact that Transformers can use low-information combinations in the input as biasing terms has been previously noted in Vision Transformers~\citet{Darcet2023VisionTN}.

\subsection{The influence of position codes}
Above we discussed the solution in the absence of a position code (i.e. NoPE). Here we look at the effect of including different types of positional encoding in the architecture.

\paragraph{AbsPE.} Transformer models with absolute position codes (AbsPE) are more expressive than those without a position code (NoPE).\footnote{In the presence of a BoS-token, a causal Transformer with NoPE can construct the AbsPE within one block. Therefore, causal Transformers with NoPE with $l+1$ blocks are at least as expressive as causal Transformers with AbsPE with $l$ blocks~\cite{Kazemnejad2023TheIO}.} We might therefore expect that the model with AbsPE should perform at least as well as those with NoPE. However, this is not the case in practice. The best models trained with NoPE are considerably better than those trained with AbsPE. Looking at the models trained with AbsPE, we see that the model finds it difficult to ignore the positional code. Instead it finds ad-hoc solutions which estimate the number of 1-tokens. For details see Sec.~\ref{app:abs}. 

\paragraph{Alibi.} Alibi uses an exponentially decaying attention bias with a different exponent for each attention head. As we only have one attention head in our model, we used different values of the exponent ranging from 1 to 0.01. When the exponent is small, Alibi attends to all tokens equally similar to NoPE. When the exponent is large, Alibi only attends locally and fails to converge to a solution with better than chance performance. More details on Alibi are given in Appendix Sec.~\ref{app:alibi}. 

\paragraph{RoPE.} This position encoding rotates different parts of the latent space with different frequencies depending on the position of the token in the sequence. The model can therefore encode information in the subspace corresponding to high frequencies with  high positional rotation or in low frequencies where the rotation is negligible. In the latter case, the behavior of the model is similar to NoPE. The best performing models trained with RoPE indeed employ this strategy and their circuits are identical to that of the best model trained with NoPE discussed above (See Sec.~\ref{app:rope}). Note that while the best solutions of NoPE and RoPE are identical in their circuits, RoPE is much more likely to find good solutions and NoPE can often fail to converge to good solutions.

\subsection{Experiments on large language models}
\label{sub:llm}

In this section, we provide preliminary results on how state-of-the-art LLMs can do on this task, with various few-shot prompting strategies.

In Table~\ref{tab:gpt4}, we show the results of the GPT-4o model on this task, with few-shot prompting, either for directly predicting the counts in the four regions, or with simple chain-of-thought (CoT) strategies that first extracts the subsequences of each region, including a variant (CoT 2) that keeps track of counting. We use 4 few-shot examples, and the evaluation is over 30 test sequences. The details of the prompts are given in Appendix~\ref{app:llms_fail}.
We see that the CoT strategies performs much better than the direct prediction strategy, and that performance degrades significantly at longer lengths. In particular, for length 512, the accuracy is much worse than the best performing architectures in Figure~\ref{fig:counting_acc_orig}.


\begin{table}[h!]
\centering
\begin{tabular}{ccc}
\toprule
\textbf{Type} & \textbf{Length} & \textbf{Accuracy} \\ 
\midrule
\multirow{3}{*}{Direct} & 128 & 19\% \\ 
 & 256 & 11\% \\ 
 & 512 & 2\% \\ \midrule
\multirow{3}{*}{CoT 1} & 128 & 62\% \\ 
 & 256 & 52\% \\ 
 & 512 & 21\% \\ \midrule
\multirow{3}{*}{CoT 2} & 128 & 94\% \\ 
 & 256 & 59\% \\ 
 & 512 & 21\% \\ 
\bottomrule
\end{tabular}
\caption{Accuracy of GPT-4o for different input sequence lengths and prompting strategy types. See also Table~\ref{tab:gpt_long} in Appendix~\ref{app:llms_fail}.}
\label{tab:gpt4}
\end{table}


\section{Conclusion}
\label{sec:conclusion}
In this work, we introduced a novel contextual counting task designed to probe the interpretability of Transformers in quantitative and scientific contexts. This task requires the model to identify specific regions within a dataset and perform accurate counting, mimicking scenarios where precise localization and subsequent computation are crucial. We conducted comprehensive theoretical and empirical analyses using both causal and non-causal Transformer architectures, exploring the influence of various positional encodings on performance and interpretability. Our findings provide valuable insights into how different positional information influences model behavior in quantitative settings. We observed that causal models outperform non-causal models, and that NoPE achieves the best performance but also exhibits the highest variance in training. We further identified several distinct solution classes with varying generalization performance, highlighting the importance of understanding the inner workings of these models.

The focus of this work was on trying to understand and interpret the nature of the solutions found in different configurations. We noted that different solution types exist, some of which do generalize out of distribution. We did not explore detail what causes a training regimen to find one solution as opposed to another and how we can improve the chances that training via SGD leads to a generalizable solution. We leave this important question to future work.

\nocite{langley00}

\bibliography{biblio}
\bibliographystyle{icml2024}

\newpage
\appendix
\onecolumn

\section{Explored configurations }
\label{app:configs}
In this section we provide a short summary of the different Transformer configurations that we analyze in this work.

\subsection{Position Codes}
We look at the influence of the following positional codes.
\paragraph{Absolute Positional Encoding (AbsPE):} As in the original Transformer architecture~\cite{Vaswani2017AttentionIA}, AbsPE denotes the global position of a token in a sequence. These can be either fixed or learned. In this work we only look at learned absolute position embedding.
\paragraph{No Positional Encoding (NoPE):}  In causal Transformers, it has been shown that even without any positional information, the model can infer absolute or relative position information~\cite{Kazemnejad2023TheIO}.
\paragraph{Rotary Positional Encoding (RoPE):} This is a commonly used positional code which encodes the positional information as a rotation of the latent space according to the position of each token~\cite{Su2021RoFormerET}.
\paragraph{Alibi:} In Alibi~\cite{Press2021TrainST}, the positional information is not appended to the latent variables but it is inserted in the attention map of the Transformer blocks by adding a negative value linearly proportional to the exponential of the distance between the query and key tokens. 


\subsection{Causal vs. Non-causal}
We explore the effect of using causal vs non-causal attention in the attention modules of the Transformer. Causal attention refers to the case where each token can only attend to itself and tokens that precede it in the sequence. In comparison, tokens in non-causal Transformers can attend to every token present in the input.

\section{LLM responses to contextual counting task}
\label{app:llms_fail}

Figures~\ref{fig:few-shot-direct} and~\ref{fig:few-shot-cot} show the few-shot prompting strategies for the ``direct'' and ``chain-of-thought (CoT)'' type of prompts mentioned in Section~\ref{sub:llm}. For CoT, we include a variant ``CoT 2'' which helps counting by keeping track of a counter. Table~\ref{tab:gpt_long} provides an extended version of Table~\ref{tab:gpt4}, including mean absolute error (MAE) and per-region metrics.

\begin{figure}[h!]
\centering
\begin{tabular}{|p{0.9\textwidth}|}
\hline
\textbf{Role: System} \\
\hline
\texttt{
You'll be provided lists of integers in the list \{0,1,2,3\}. Your goal is to return a list of 4 elements, where each corresponds to the number of 1 tokens in each of the 4 regions surrounded by start (2) and end (3) tokens. Use a consistent style.
} \\
\hline
\end{tabular}

\vspace{1em}

\begin{tabular}{|p{0.9\textwidth}|}
\hline
\textbf{Role: User} \\
\hline
\texttt{
[0, 1, 1, 0, 0, 1, 1, 0, 1, 1, 2, 1, 1, 3, 0, 2, 0, 0, 1, 1, 1, 1, 1, 3, 0, 0, 0, 0, 0, 1, 1, 1, 1, 0, 1, 1, 1, 1, 1, 1, 2, 1, 0, 1, 1, 1, 1, 0, 1, 0, 1, 0, 1, 0, 0, 0, 1, 1, 1, 1, 0, 0, 1, 1, 0, 0, 1, 0, 0, 1, 1, 0, 1, 0, 3, 0, 0, 1, 2, 0, 1, 1, 1, 1, 1, 0, 1, 0, 0, 1, 1, 0, 1, 0, 0, 0, 0, 1, 0, 1, 0, 3, 1, 0, 0, 0, 1, 1, 1, 1, 0, 0, 1, 1, 0, 1, 0, 0, 1, 0, 0, 0, 1, 0, 0, 0, 1, 1]
} \\
\hline
\end{tabular}

\vspace{1em}

\begin{tabular}{|p{0.9\textwidth}|}
\hline
\textbf{Role: Assistant} \\
\hline
\texttt{[2, 5, 18, 11]} \\
\hline
\end{tabular}

\caption{Prompt and first few-shot example for ``Direct''}
\label{fig:few-shot-direct}
\end{figure}

\begin{figure}[h!]
\centering
\begin{tabular}{|p{0.9\textwidth}|}
\hline
\textbf{Role: System} \\
\hline
\texttt{
You'll be provided lists of integer tokens in the set \{`0`,`1`,`2`,`3`\}. Your goal is to identify the four regions surrounded by `2` and `3` tokens, and count the number of `1` tokens in each region. Use a consistent style.
} \\
\hline
\end{tabular}

\vspace{1em}

\begin{tabular}{|p{0.9\textwidth}|}
\hline
\textbf{Role: User} \\
\hline
\texttt{
[0, 1, 1, 0, 0, 1, 1, 0, 1, 1, 2, 1, 1, 3, 0, 2, 0, 0, 1, 1, 1, 1, 1, 3, 0, 0, 0, 0, 0, 1, 1, 1, 1, 0, 1, 1, 1, 1, 1, 1, 2, 1, 0, 1, 1, 1, 1, 0, 1, 0, 1, 0, 1, 0, 0, 0, 1, 1, 1, 1, 0, 0, 1, 1, 0, 0, 1, 0, 0, 1, 1, 0, 1, 0, 3, 0, 0, 1, 2, 0, 1, 1, 1, 1, 1, 0, 1, 0, 0, 1, 1, 0, 1, 0, 0, 0, 0, 1, 0, 1, 0, 3, 1, 0, 0, 0, 1, 1, 1, 1, 0, 0, 1, 1, 0, 1, 0, 0, 1, 0, 0, 0, 1, 0, 0, 0, 1, 1]
} \\
\hline
\end{tabular}

\vspace{1em}

\begin{tabular}{|p{0.9\textwidth}|}
\hline
\textbf{Role: Assistant} \\
\hline
\texttt{The four regions with the corresponding count of `1` tokens are:} \\
\texttt{* [2, 1, 1, 3]: 2} \\
\texttt{* [2, 0, 0, 1, 1, 1, 1, 1, 3]: 5} \\
\texttt{* [2, 1, 0, 1, 1, 1, 1, 0, 1, 0, 1, 0, 1, 0, 0, 0, 1, 1, 1, 1, 0, 0, 1, 1, 0, 0, 1, 0, 0, 1, 1, 0, 1, 0, 3]: 18} \\
\texttt{* [2, 0, 1, 1, 1, 1, 1, 0, 1, 0, 0, 1, 1, 0, 1, 0, 0, 0, 0, 1, 0, 1, 0, 3]: 11} \\
\hline
\end{tabular}

\caption{Prompt and first few-shot example for chain-of-thought ``CoT 1''}
\label{fig:few-shot-cot}
\end{figure}

\begin{figure}[h!]
\centering
\begin{tabular}{|p{0.9\textwidth}|}
\hline
\textbf{Role: System} \\
\hline
\texttt{You'll be provided lists of integer tokens in the set \{`0`,`1`,`2`,`3`\}. Your goal is to identify the four regions surrounded by `2` and `3` tokens, and count the number of `1` tokens in each region. When printing each sequence, you can use parentheses to keep track of counting. Use a consistent style.} \\
\hline
\end{tabular}

\vspace{1em}

\begin{tabular}{|p{0.9\textwidth}|}
\hline
\textbf{Role: User} \\
\hline
\texttt{[0, 1, 1, 0, 0, 1, 1, 0, 1, 1, 2, 1, 1, 3, 0, 2, 0, 0, 1, 1, 1, 1, 1, 3, 0, 0, 0, 0, 0, 1, 1, 1, 1, 0, 1, 1, 1, 1, 1, 1, 2, 1, 0, 1, 1, 1, 1, 0, 1, 0, 1, 0, 1, 0, 0, 0, 1, 1, 1, 1, 0, 0, 1, 1, 0, 0, 1, 0, 0, 1, 1, 0, 1, 0, 3, 0, 0, 1, 2, 0, 1, 1, 1, 1, 1, 0, 1, 0, 0, 1, 1, 0, 1, 0, 0, 0, 0, 1, 0, 1, 0, 3, 1, 0, 0, 0, 1, 1, 1, 1, 0, 0, 1, 1, 0, 1, 0, 0, 1, 0, 0, 0, 1, 0, 0, 0, 1, 1]} \\
\hline
\end{tabular}

\vspace{1em}

\begin{tabular}{|p{0.9\textwidth}|}
\hline
\textbf{Role: Assistant} \\
\hline
\texttt{The four regions with the corresponding count of `1` tokens are:} \\
\texttt{* [2, 1 (1), 1 (2), 3]: 2} \\
\texttt{* [2, 0 (0), 0 (0), 1 (1), 1 (2), 1 (3), 1 (4), 1 (5), 3]: 5} \\
\texttt{* [2, 1 (1), 0 (1), 1 (2), 1 (3), 1 (4), 1 (5), 0 (5), 1 (6), 0 (6), 1 (7), 0 (7), 1 (8), 0 (8), 0 (8), 0 (8), 1 (9), 1 (10), 1 (11), 1 (12), 0 (12), 0 (12), 1 (13), 1 (14), 0 (14), 0 (14), 1 (15), 0 (15), 0 (15), 1 (16), 1 (17), 0 (17), 1 (18), 0 (18), 3]: 18} \\
\texttt{* [2, 0 (0), 1 (1), 1 (2), 1 (3), 1 (4), 1 (5), 0 (5), 1 (6), 0 (6), 0 (6), 1 (7), 1 (8), 0 (8), 1 (9), 0 (9), 0 (9), 0 (9), 0 (9), 1 (10), 0 (10), 1 (11), 0 (11), 3]: 11} \\
\hline
\end{tabular}

\caption{Prompt and first few-shot example for ``CoT 2''}
\label{fig:few-shot-cot2}
\end{figure}

\begin{table*}[h!]
\centering
\begin{tabular}{cccccc}
\toprule
\textbf{Type} & \textbf{Length} & \textbf{Accuracy} & \textbf{Region Accuracies} & \textbf{MAE} & \textbf{Region MAE} \\ 
\midrule
\multirow{3}{*}{Direct} & 128 & 0.19 & [0.27, 0.23, 0.17, 0.1] & 2.90 & [2.6, 2.1, 3.57, 3.33] \\ 
 & 256 & 0.11 & [0.2, 0.03, 0.2, 0.0] & 11.05 & [7.23, 8.8, 8.67, 19.5] \\ 
 & 512 & 0.017 & [0.033, 0.0, 0.033, 0.0] & 23.5 & [27.3, 23.2, 19.7, 23.8] \\ \midrule
\multirow{3}{*}{CoT 1} & 128 & 0.62 & [0.67, 0.63, 0.47, 0.7] & 0.62 & [0.67, 0.57, 0.9, 0.37] \\ 
 & 256 & 0.52 & [0.63, 0.6, 0.43, 0.4] & 1.13 & [0.77, 0.57, 1.9, 1.3] \\ 
 & 512 & 0.21 & [0.33, 0.29, 0.083, 0.12] & 9.68 & [2.79, 5.37, 12.5, 18.0] \\ \midrule
\multirow{3}{*}{CoT 2} & 128 & 0.94 & [0.97, 0.93, 0.97, 0.90] & 0.22 & [0.07, 0.21, 0.03, 0.59] \\ 
 & 256 & 0.59 & [0.79, 0.55, 0.48, 0.52] & 3.41 & [0.48, 3.0, 6.86, 3.28] \\ 
 & 512 & 0.21 & [0.67, 0.0, 0.17, 0.0] & 20.13 & [2.5, 22.0, 29.83, 26.17] \\ \bottomrule
\end{tabular}
\caption{Comparison of Accuracy and Mean Absolute Error for different input sequence lengths and prompt types. Extended version of Table~\ref{tab:gpt4}, including per-region accuracy and mean absolute error. CoT denotes the chain-of-thought type of prompt.}
\label{tab:gpt_long}
\end{table*}

\section{Proof sketches}
\label{app:proofs}

\paragraph{Proposition 3.1}\textit{
(informal) If the regional contextual position information is linearly decodable from the latent representation of the tokens at some layer of a Transformer, the Contextual Counting task can be solved with a single additional layer.} 

This statement is a corollary of the construction of the absolute position code from NoPE in Theorem 1 of~\cite{Kazemnejad2023TheIO}. There, it is shown that by attending to every prior position, every token can infer the total number of prior tokens. By extension, by attending to only the 1-tokens, in a cross-attention module, we can similarly infer the total number of 1-tokens. 

If the regional contextual position information is available, the cross-attention module can attend only to a specific region. The output of this cross attention module will be proportional to $n v_1 + m v_0 + v_[ + v_]$, where $n$ is the number of 1-tokens (i.e. the quantity we are after), $m$ is the number of 0-tokens in the region, and $v_0$, $v_1$, $v_[$, and $v_]$ are the value vectors coming from each token type. If we set $v_0 = v_]=0$, $N$ is encoded in the ratio of the magnitude of  $v_1$ and $v_[$. The construction of this section is exactly analogous to that of~\cite{Kazemnejad2023TheIO}, with the [-token playing the role of BoS-token of~\cite{Kazemnejad2023TheIO}. 

\paragraph{Proposition 3.2}\textit{
(informal) A causal Transformer with a single layer and no position encoding (NoPE) can infer the regional contextual position for arbitrary number of regions and arbitrary sequence lengths.}
This proposition is also a corollary of the construction of the absolute position code from NoPE in Theorem 1 of~\cite{Kazemnejad2023TheIO}. In this case, each token can attend equally to previous delimiter tokens arriving at a latent variable which is given by$(n_[ v_[ + n_] v_])/(n_[  + n_])$. When a token is not between delimiters, that is in an inter-region, $n_[ = n_]$ and the latent reduces to $( v_[ + v_])/2$. For the regions of interest, $n_[ = n_]+1$ implying that we are between two delimiters. In this case, the desired quantity is encoded in the angle between the latent and $n_[$. This behavior is exactly seen in Fig.~\ref{fig:counting_loss_nomlp_bestsol_ones_postatt}, where all the different inter-regions have collapsed onto the same point.

\paragraph{Proposition 3.3}\textit{
    (informal) A non-causal Transformer with no position code and a permutation invariant output head cannot solve the Contextual Counting task.}
A non-causal Transformer with no position code is permutation equivariant. If the output of this model is extracted in a permutation invariant way (e.g. averaging or cross-attention),  the output of the model will be permutation invariant. As the contextual counting task is sensitive to the ordering of the elements of the sequence, it cannot be solved by a permutation invariant network.  

\paragraph{Proposition 3.4}\textit{
    (informal) To exactly emulate a causal attention profile, a non-causal Transformer with Absolute Position code would need an embedding space that has dimension at least as large as the length of the sequence.  }
\begin{proof}
    Proof by contradiction. Assume a sequence of length $T$ and embedding dimension of the keys given by $d<T$. This implies that the key vector for the $T$'th token is a linear combination of the other key vectors $k_T = \sum_{t<T} c_t k_t$. Taking an inner-product with the query vector of the $T$'th token we have $k_T\cdot q_T = \sum_{t<T} c_t k_t\cdot q_T$. If we want to exactly emulate a causal attention profile we need  $k_T\cdot q_T=1$ and $k_t\cdot q_T=0$ if $t<T$. Therefore, the previous sum turns to $1 = k_T\cdot q_T = \sum_{t<T} c_t k_t\cdot q_T=0$, which is a contradiction.
\end{proof}

\section{Other model details}

\subsection{Models trained with Absolute Position Code}
\label{app:abs}

\begin{figure}
    \centering
    \includegraphics[width=0.45\textwidth]{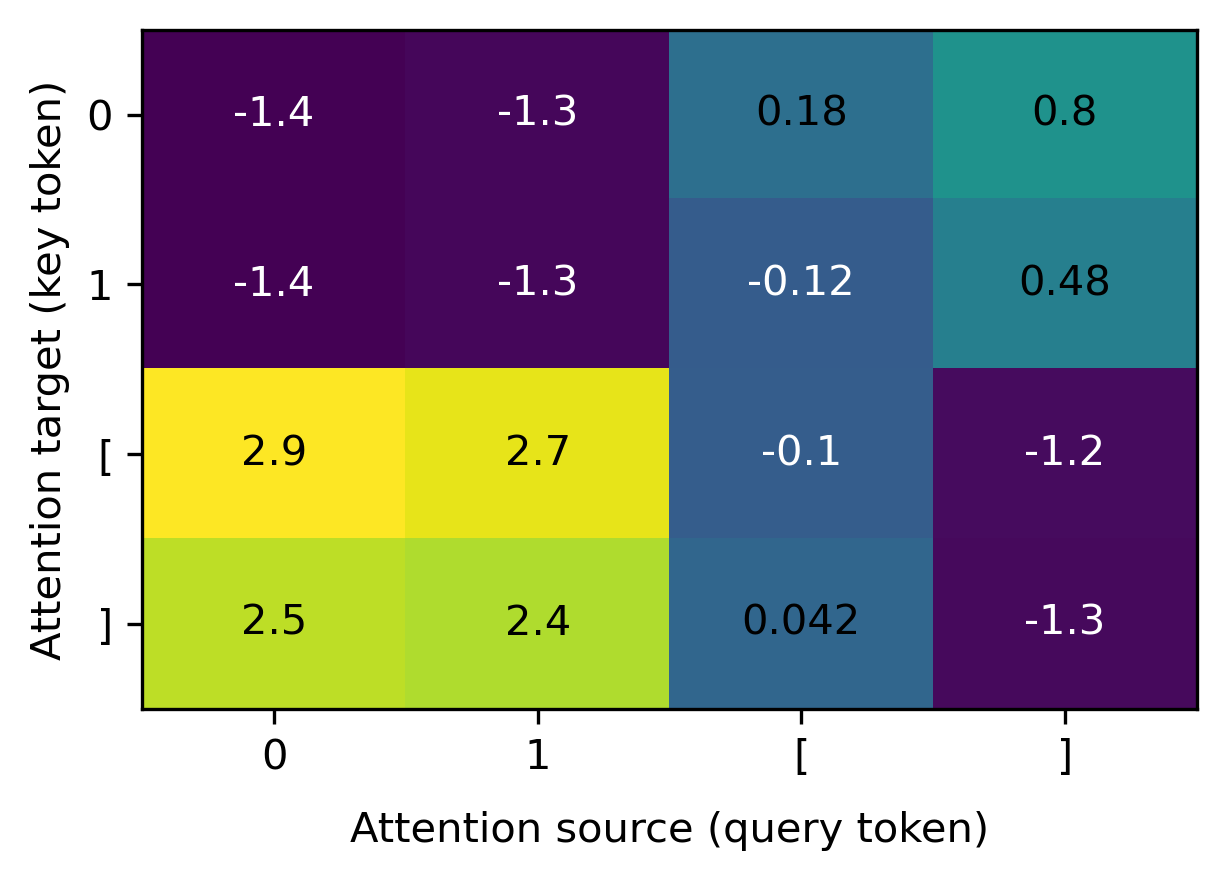}
    \vspace{-4pt}
    \caption{The contextual attention pattern of encoder the best model trained with Absolute Position code. This plot shows only the attention weights coming from the context, removing the position dependence.}
    \label{fig:counting_loss_abs_bestsol_enc_att}
\end{figure}

In this section we provide some further details of the best performing model trained with AbsPE. The contextual attention in the encoder layer resembles that of the previous solutions with the 1-tokens attending primarily to the delimiters (Fig.~\ref{fig:counting_loss_abs_bestsol_enc_att}).

However, in this case, context is not the only contributor to the attention and the self-attention module can choose to attend to the location information as well (Fig.~\ref{fig:counting_loss_abs_bestsol_enc_att_loc}). Here, again for clarity we only show the attention weights for an input of length 100. We see that the model pays preferential attention to specific locations in the sequence, even though there are no special positions in the sequence a-priori. 
\begin{figure}
    \centering
    \includegraphics[width=0.55\textwidth]{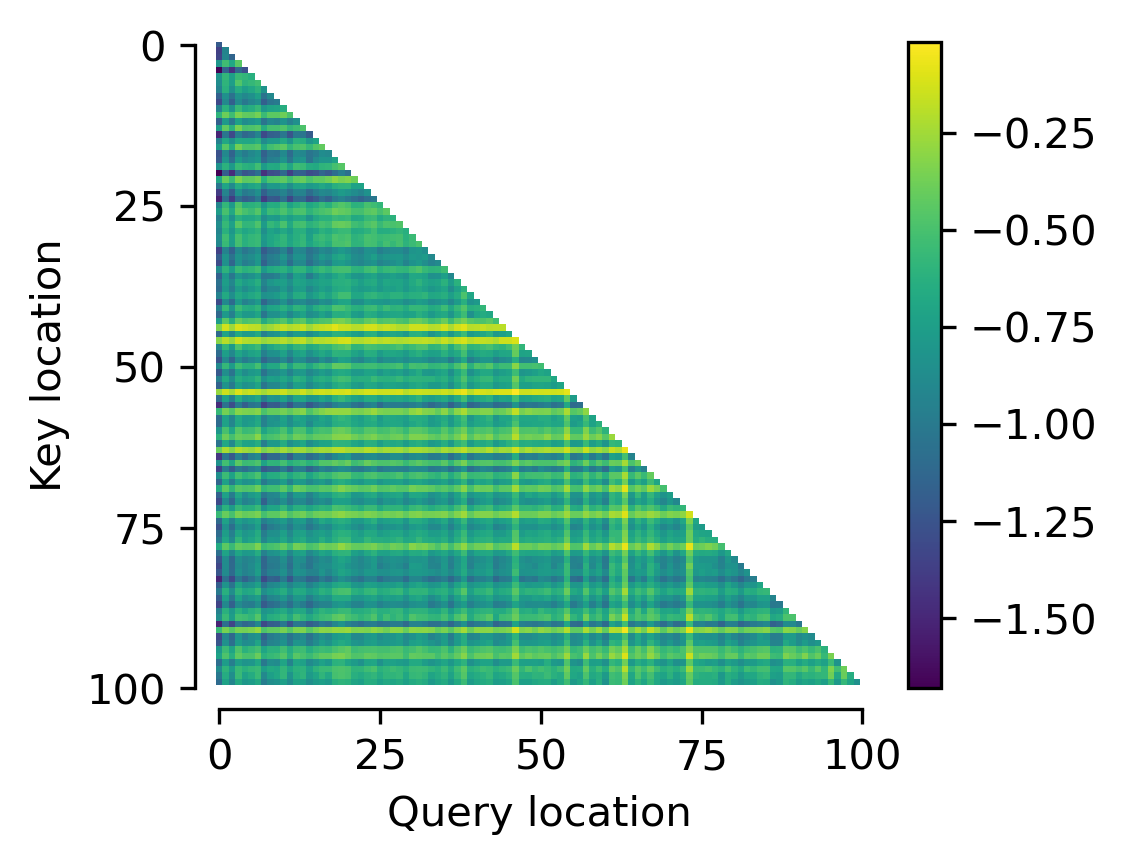}
    \vspace{-4pt}
    \caption{The positional attention pattern of encoder the best model trained with Absolute Position code. This plot shows only the attention weights coming from the location, removing the contextual dependence.}
    \label{fig:counting_loss_abs_bestsol_enc_att_loc}
\end{figure}

This positional dependence is also present in the attention pattern of the decoder module (Fig.~\ref{fig:counting_loss_abs_bestsol_enc_att_loc}). In this case, interpreting exactly what algorithm the model is implementing is more challenging as tokens at specific positions perform all carry out different computations. The result of this ad-hoc solution is that it severely under-performs the other solutions discussed.
\begin{figure}
    \centering
    \includegraphics[width=0.55\textwidth]{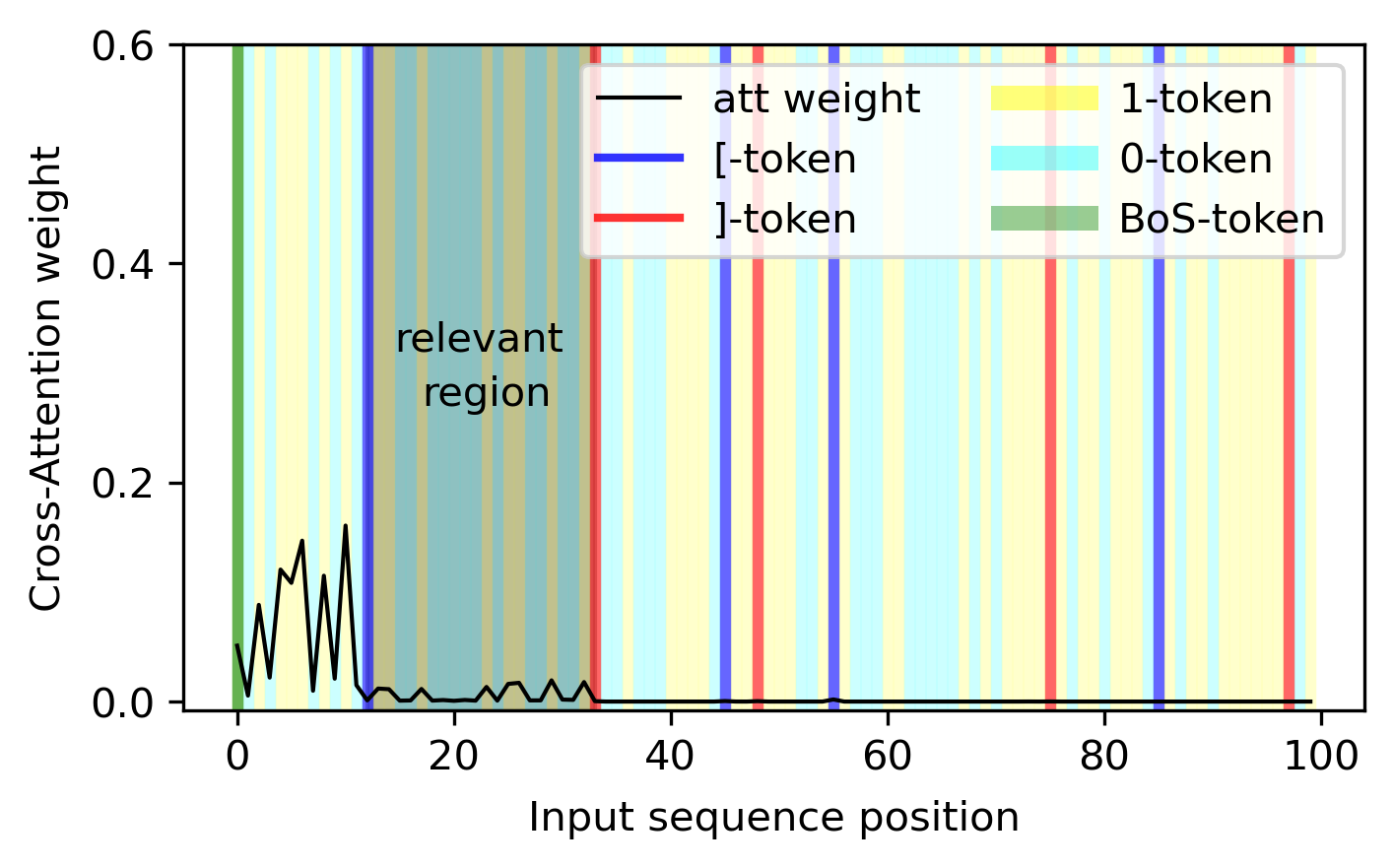}
    \includegraphics[width=0.55\textwidth]{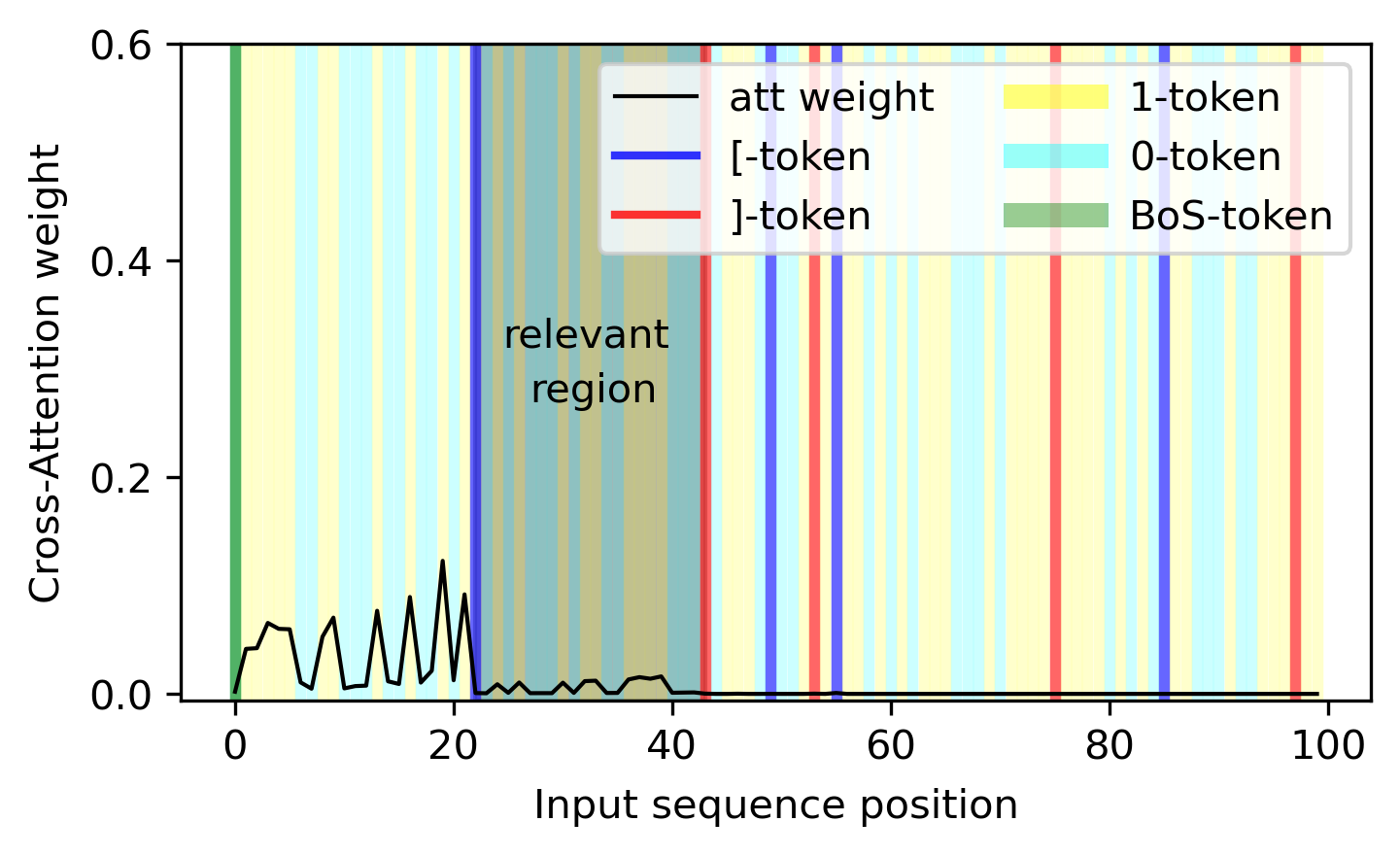}
    \vspace{-4pt}
    \caption{Two examples of the attention weights of the cross-attention module of the decoder in the model trained with AbsPE. The shaded regions denote the identity of the tokens at the indicated position. As with other cases, the model was trained on length 512 input sequences but for demonstration, we show a sample of length 100.}
    \label{fig:counting_loss_abs_bestsol_enc_att_loc}
\end{figure}

\subsection{Model trained with decoder MLP}
\label{app:MLP}
\begin{figure}
    \centering
    \includegraphics[width=0.45\textwidth]{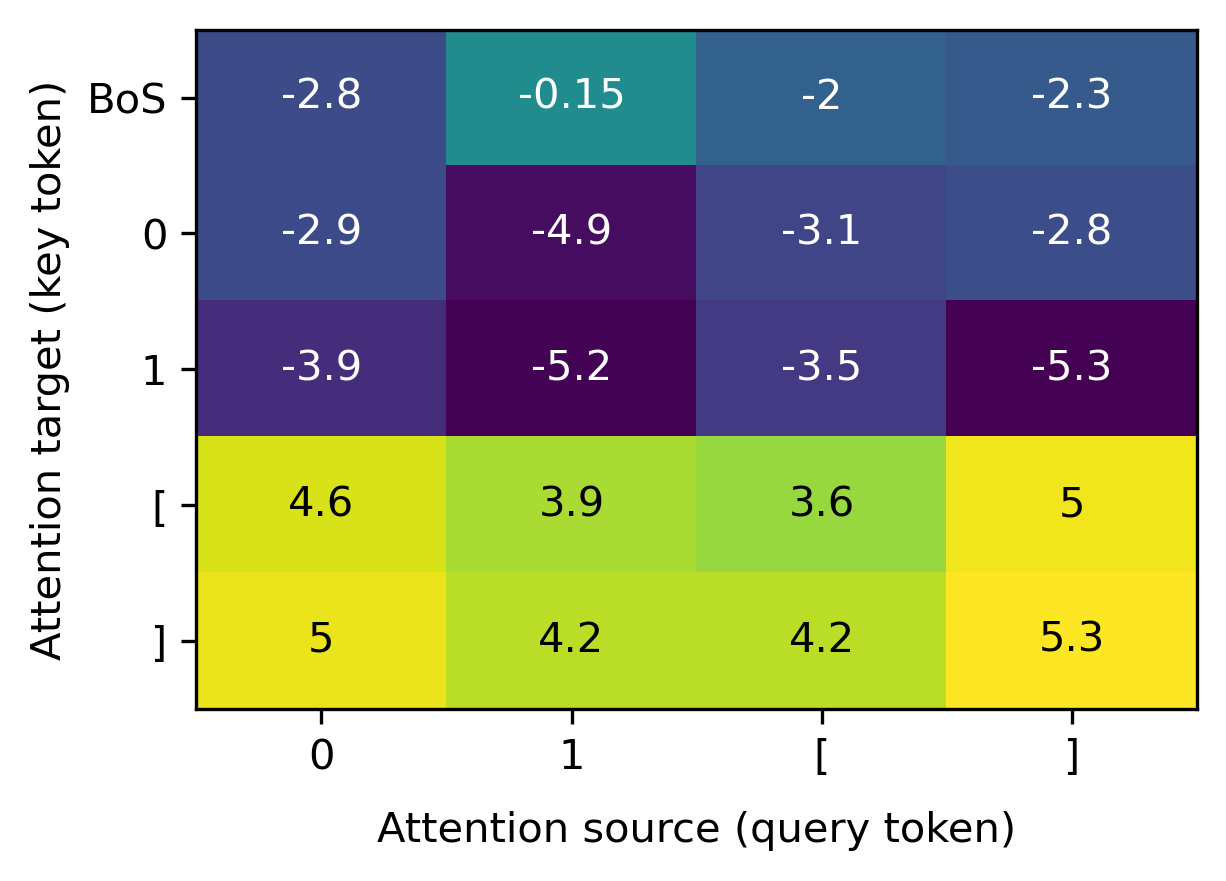}
    \vspace{-4pt}
    \caption{The encoder attention pattern of the best model trained with a decoder MLP.}
    \label{fig:counting_loss_mlp_bestsol_enc_att}
\end{figure}

In the main body of the paper we focused on models trained without an MLP layer in the decoder block. The linearity that follows from this choice made it possible to look at the contribution of each token. However, we see that the attention patterns for different solutions remain similar to that of the previous cases. For example, in the encoder of most successful models, the 1-tokens attend the previous delimiter tokens to find their own contextual position (Fig.~\ref{fig:counting_loss_mlp_bestsol_enc_att}). And this information is used in the decoder so that in the cross-attention layer, the module  attends to the 1-tokens that are in the relevant region (Fig.~\ref{fig:counting_loss_nomlp_bestsol_crossatt}). In this case, the learned model ignores the BoS sequence in favor of the ]-tokens which it uses as a bias. 

In this way we can verify that the overall algorithm implemented in models with MLP are similar to that of the models implemented with MLP.

\begin{figure}
    \centering
    \includegraphics[width=0.55\columnwidth]{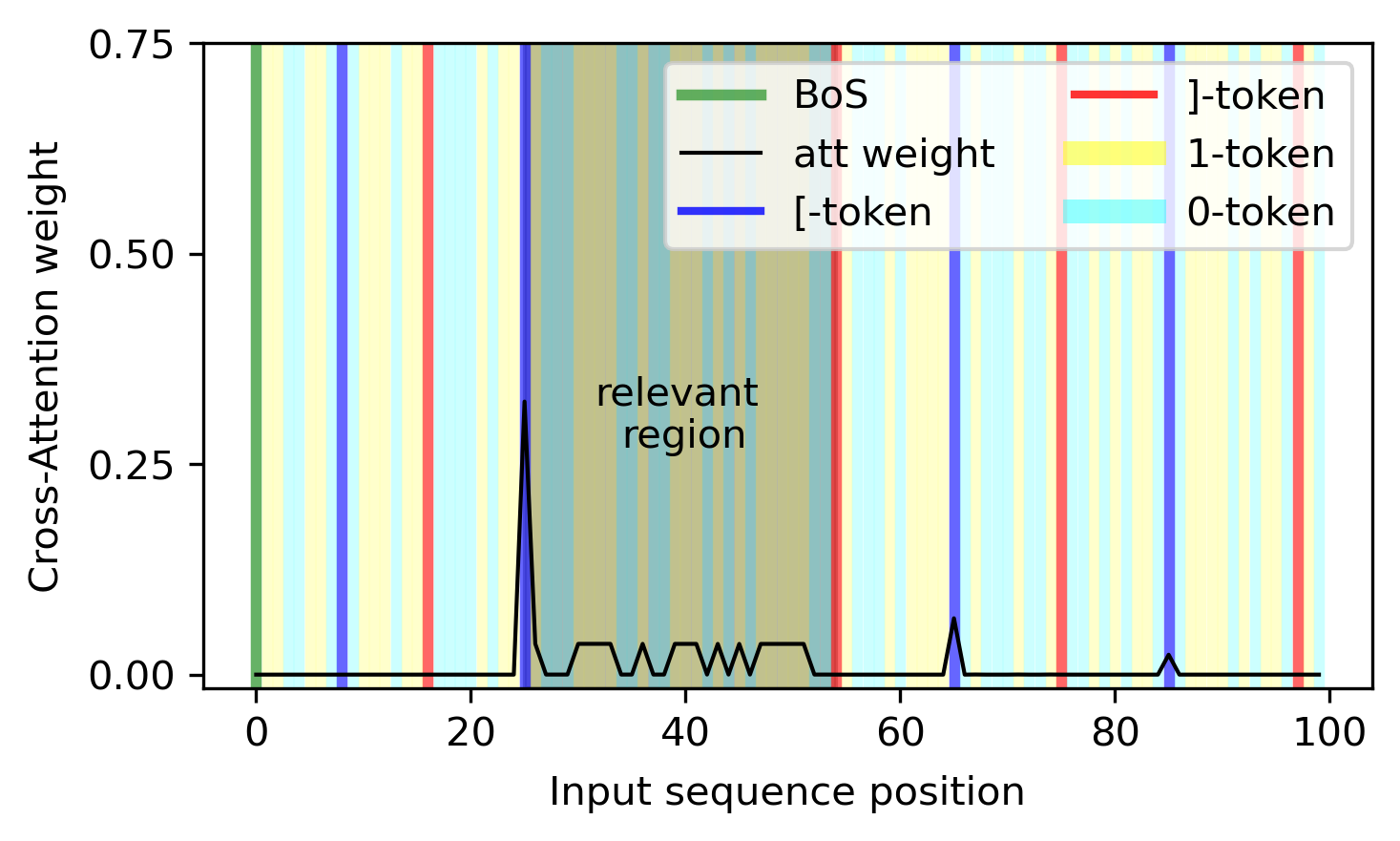}
    \vspace{-4pt}
    \caption{The attention weights of the cross-attention module of the decoder trained with a decoder MLP. The shaded regions denote the identity of the tokens at the indicated position. As with other cases, the model was trained on length 512 input sequences but for demonstration, we show a sample of length 100.}
    \label{fig:counting_loss_nomlp_bestsol_crossatt}
\end{figure}

\subsection{Alibi}
\label{app:alibi}

\begin{figure}
    \centering
    \includegraphics[width=0.5\textwidth]{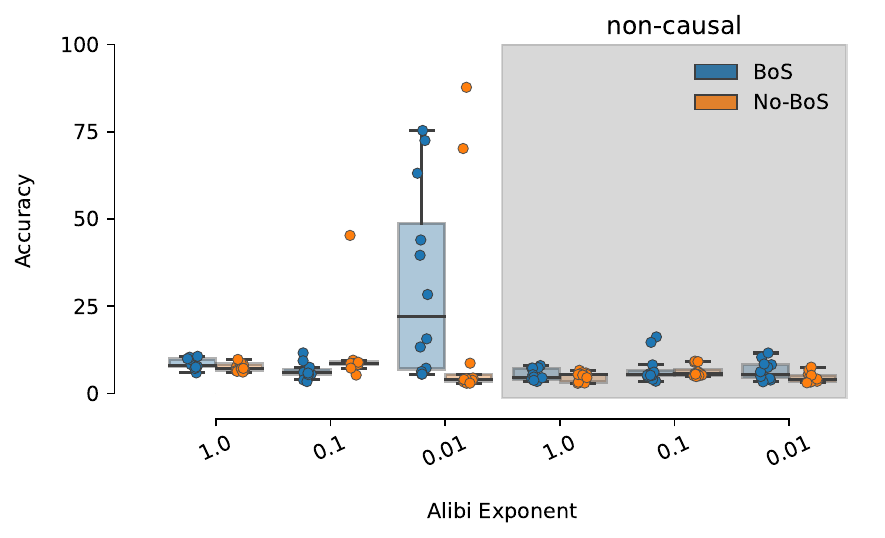}
    \vspace{-10pt}
    \caption{Model Accuracy on the Contextual Counting task on models trained with Alibi. }
    \label{fig:alibi_orig}
\end{figure}

\begin{figure}
    \centering
    \includegraphics[width=0.5\textwidth]{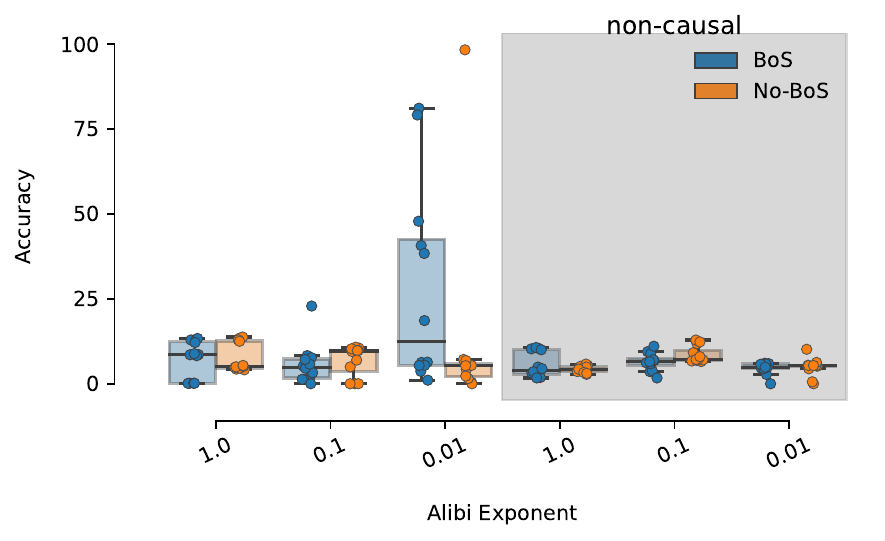}
    \vspace{-10pt}
    \caption{Generalization performance on test samples with shorter sequences on models trained with Alibi. }
    \label{fig:alibi_shorter}
\end{figure}

\begin{figure}
    \centering
    \includegraphics[width=0.5\textwidth]{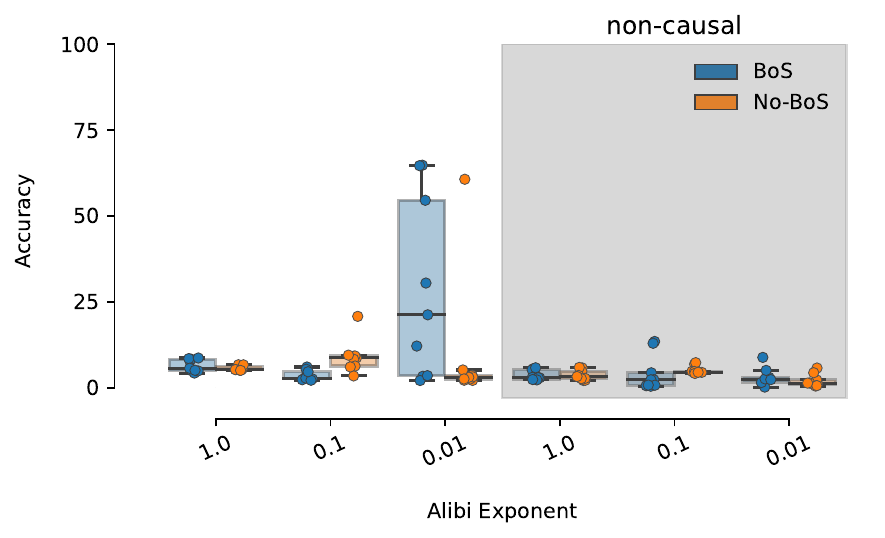}
    \vspace{-10pt}
    \caption{Generalization performance on test samples with fewer regions on models trained with Alibi. }
    \label{fig:alibi_3bounds}
\end{figure}

The performance of models trained with Alibi are given in Figs.~\ref{fig:alibi_orig},~\ref{fig:alibi_shorter}, and~\ref{fig:alibi_3bounds}. We see that the performance of Alibi gets better as the exponent gets smaller. This can be understood as the limit of the exponent going to zero in Alibi is the same as NoPE which was discussed at length in the main body of the manuscript. We also see here that short-range attention, as implemented by Alibi with higher values of the exponent cannot solve this task with a single head. 

\subsection{Rope}
\label{app:rope}

\begin{figure}
    \centering
    \includegraphics[width=0.5\textwidth]{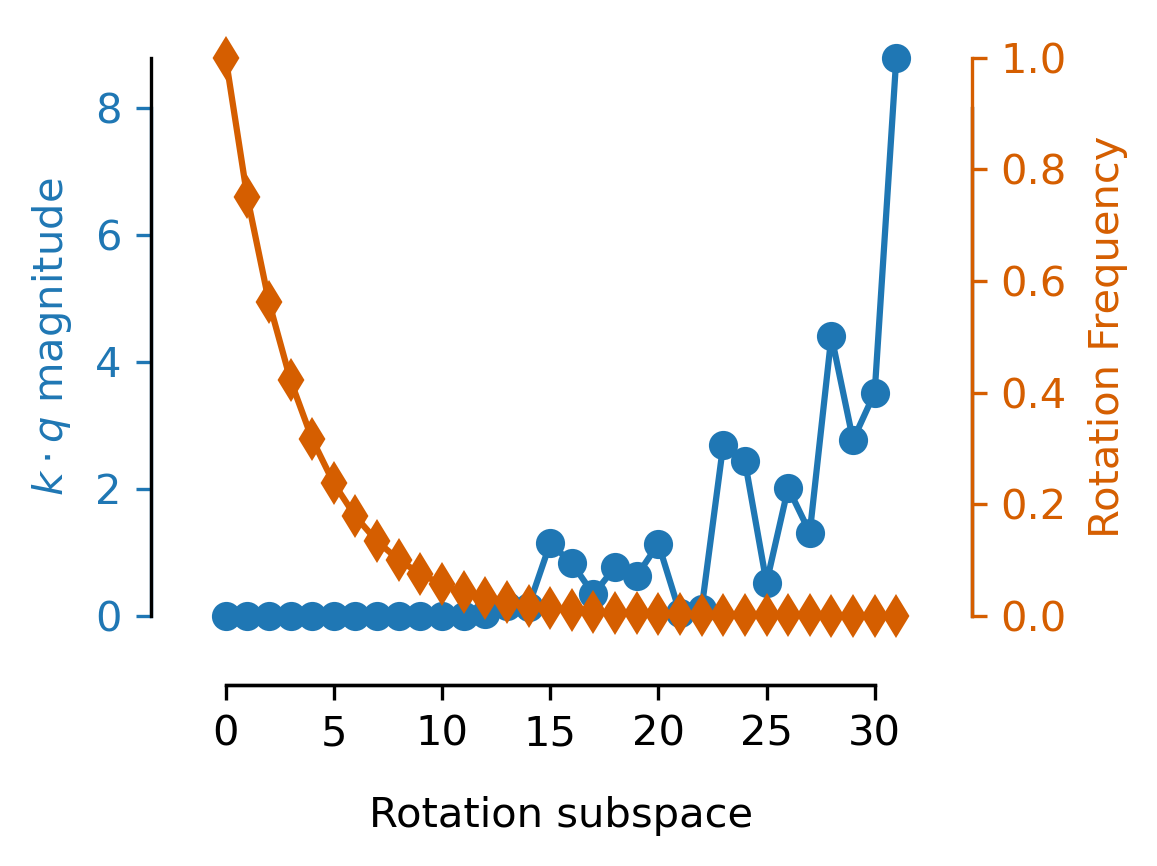}
    \caption{(Blue) The magnitude of the encoder attention inner product decomposed into the different rotational subspaces. (Orange) The rotation frequency of the subspace. }
    \label{fig:rope_subspace}
\end{figure}

Here, we look at the best model trained with RoPE. As mentioned in the main body of the manuscript, RoPE has the flexibility of encoding information in subspaces that rotate with different frequencies. Figure~\ref{fig:rope_subspace} shows the contribution of the different rotational subspaces to attention map the encoder layer. We see that the attention weights are primarily coming from subspaces where the rotation frequency is smallest. 

This implies that the model is choosing to ignore the positional information and is effectively implementing the same strategy as the solutions discussed for NoPE. Specifically, the encoder and decoder attention patterns are similar to the solution described previously.


\clearpage
\end{document}